\documentclass{article}
\usepackage{enumitem}
\usepackage{caption}
\usepackage{bbm}
\usepackage{threeparttable}
\usepackage{wrapfig}
\usepackage[rflt]{floatflt}

\usepackage[final, nonatbib]{neurips_2021}
\usepackage[utf8]{inputenc} 
\usepackage[T1]{fontenc}    
\usepackage{hyperref}       
\usepackage{url}            
\usepackage{booktabs}       
\usepackage{amsfonts}       
\usepackage{nicefrac}       
\usepackage{microtype}      
\usepackage{todonotes}
\usepackage{algorithm,algorithmic} 

\usepackage{caption}
\usepackage{subcaption}
\usepackage{wrapfig}
\usepackage{booktabs}
\usepackage[compact]{titlesec}
\titlespacing*{\section}
{0pt}{1ex}{0.7ex}
\titlespacing*{\subsection}
{0pt}{0.6ex}{0.3ex}

\usepackage{amsmath,amssymb, amsthm}

\newcommand{\beq}{\begin{equation}}
	\newcommand{\eeq}{\end{equation}}
\newcommand{\beqa}{\begin{eqnarray}}
	\newcommand{\eeqa}{\end{eqnarray}}
\newcommand{\beqan}{\begin{eqnarray*}}
	\newcommand{\eeqan}{\end{eqnarray*}}

\newtheorem{theorem}{Theorem}
\newtheorem{lemma}{Lemma}

\newtheorem{assumption}{Assumption}

\newcommand{\argmax}{\mathop{\mathrm{argmax}}}

\newcommand{\norm}[1]{\left\|#1\right\|}

\newcommand{\ip}[2]{\left\langle #1,#2 \right\rangle}

\newcommand{\prob}[1]{\mathbb{P}\left[{#1}\right]}

\newcommand{\expectSup}[2]{\mathbb{E}^{#1}\left[{#2}\right]}
\newcommand{\probSup}[2]{\mathbb{P}^{#1}\left[{#2}\right]}
\newcommand{\expectInf}[1]{\expectSup{(\infty)}{#1}}
\newcommand{\probInf}[1]{\probSup{(\infty)}{#1}}
\newcommand{\probEq}[1]{\tilde{\mathbb{P}}\left[{#1}\right]}
\newcommand{\ThetaOne}{\Theta^{(1)}}
\newcommand{\SizeThetaOne}{\left| \ThetaOne \right|}

\newcommand{\given}{\; \big\vert \;}
\newcommand{\ignore}[1]{}
\newcommand{\shortver}[1]{}
\newcommand{\appropto}{\mathrel{\vcenter{
			\offinterlineskip\halign{\hfil$##$\cr
				\propto\cr\noalign{\kern2pt}\sim\cr\noalign{\kern-2pt}}}}}

\newcommand{\kl}[2]{D\left(#1 || #2 \right)}
\newcommand{\klshort}[3]{D_{{#1}, {#2}}\left( #3 \right)}
\newcommand{\klavg}[2]{\bar{D}_{{#1}, {#2}}}
\newcommand{\indic}[1]{\mathbf{1}\left\{ #1 \right\}}

\newcommand{\Pobs}[3]{\mathbb{P}_{#1}\left[ #2 | #3 \right]}

\newcommand{\egcd}{$\epsilon$-GCD}

\newcommand{\ExploreQueue}{Q^{(1)}}
\newcommand{\ExploitQueue}{Q^{(2)}}
\newcommand{\OptAction}[1]{a^*({#1})}
\newcommand{\AlldataQueue}{\ExploreQueue}
\newcommand{\probnew}[1]{\mathbb{P}^{(\nu, \theta^*)}_{Q} \left[ #1 \right] }
\newcommand{\expectnew}[1]{\mathbb{E}^{(\nu, \theta^*)}_{Q} \left[ #1 \right] }

\usepackage{todonotes} 

\usepackage[
backend=biber,
style=alphabetic,
useprefix=true,
sorting=anyt,
maxbibnames=99
]{biblatex}
\title{Bandit Quickest Changepoint Detection}

\author{%
  Aditya Gopalan
\\
  Electrical Communication Engineering\\
Indian Institute of Science, Bengaluru 560012\\
\texttt{aditya@iisc.ac.in}\\
   \And
   Braghadeesh Lakshminarayanan\\
   Electrical Communication Engineering\\
Indian Institute of Science, Bengaluru 560012\\
\texttt{braghadeesh94@gmail.com}\\
   \AND
   Venkatesh Saligrama\\
   Electrical and Computer Engineering\\
   Boston University, Boston, MA 02215\\
   \texttt{srv@bu.edu} \\
}

\begin{document}

\maketitle

\begin{abstract}

Many industrial and security applications employ a suite of sensors for detecting abrupt changes in temporal behavior patterns. These abrupt changes typically manifest locally, rendering only a small subset of sensors informative. Continuous monitoring of every sensor can be expensive due to resource constraints, and serves as a motivation for the bandit quickest changepoint detection problem, where sensing actions (or sensors) are sequentially chosen, and only measurements corresponding to chosen actions are observed. We derive an information-theoretic lower bound on the detection delay for a general class of finitely parameterized probability distributions. We then propose a computationally efficient online sensing scheme, which seamlessly balances the need for exploration of different sensing options with exploitation of querying informative actions. We derive expected delay bounds for the proposed scheme and show that these bounds match our information-theoretic lower bounds at low false alarm rates, establishing optimality of the proposed method. We then perform a number of experiments on synthetic and real datasets demonstrating the effectiveness of our proposed method.
\end{abstract}
\section{Introduction}
\label{sec:intro}
We propose a framework for bandit\footnote{By `bandit' we generally mean adaptive sampling with partial information.} quickest change detection (BQCD). Specifically, we are given a multi-dimensional online data stream, which can be sequentially probed by means of actions belonging to an action set (for instance, only a few among all of the data stream components can be observed at any time, or a linear combination can be acquired). The online data stream can exhibit abrupt {\it statistical} changes, at any time, and only among a few arbitrary components. Our task is to sequentially probe the data stream by adaptively choosing valid actions based on past bandit\footnote{This is different from classical quickest change detection~\cite{tartakovsky2014sequential}, where all of the multiple streams are observed at each time, and the only adaptive decision is to declare whether or not a change has happened.} observations associated with past actions. Our objective is to detect a change, if it has happened, with minimum detection delay.



\begin{wrapfigure}[38]{r}{0.5\textwidth}
\vspace{-8pt}
    \includegraphics[width=0.47\textwidth]{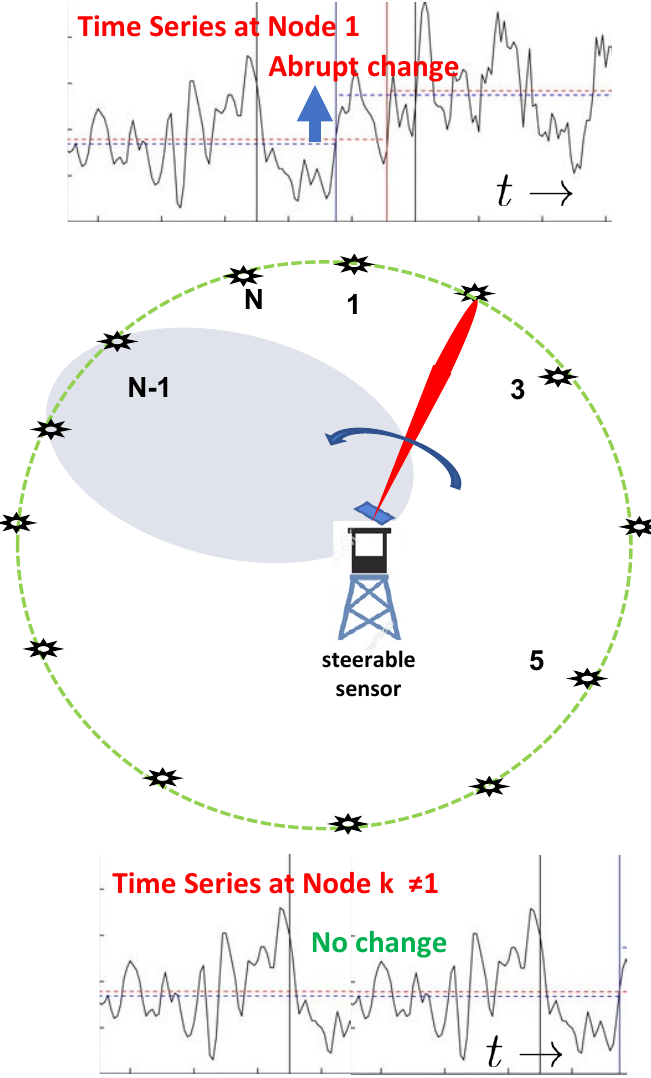}
  \caption{\footnotesize Example BQCD Scenario. There are $N$ spatial-locations that are monitored. The steerable sensor is capable of rapidly switching modes--grey and red beams as shown--and collect aggregated signal from multiple locations at a low signal-to-noise ratio (SNR) or focus attention narrowly on a single location at high SNR. Locations can exhibit abrupt change, which manifests as the depicted time-series (see top). Due to resource constraints, at any time, only a limited set of locations can be observed with high SNR. BQCD aims to learn a policy to adaptively steer and switch modes to different locations and detect abrupt changes if any, quickly.}
  \label{fi:bqcd}
  \vspace{-24pt}
\end{wrapfigure}

{\it Example Scenarios.} Surveillance systems~\cite{hero_smg} are equipped with a suite of sensors that can be switched and steered to focus attention on any target or location over a physical landscape (see Fig.~\ref{fi:bqcd}) to detect abrupt changes at any location.
On the other hand, sensor suites are resource limited, and only a limited subset, among all the locations, can be probed at any time. As such, we face a fundamental {\it dilemma}: Focusing attention at any one location can compromise change detection at other locations. 
Although we describe a scenario in surveillance, the problem of BQCD is general and arises in intrusion detection~\cite{Bass99multisensordata}, social networks~\cite{Viswanath2014TowardsDA}, disease outbreaks and epidemics \cite{qian14,Yang14473}, manufacturing processes~\cite{purohit2019,ding06}, energy limited sensor networks\cite{osborne,ermis_saligrama_2010}, and vital health monitoring \cite{Villarroel17}.

In this paper, we derive a fundamental characterization for the delay performance of BQCD, and make explicit, the fundamental tradeoff between delay and false-alarm in the low-false alarm regime.
We specifically make the following contributions. 

\noindent {\bf Information-theoretic Lower Bound.} We prove a lower bound on the expected detection delay that any BQCD algorithm must suffer at a fixed false alarm rate. The lower bound exhibits a fundamental tradeoff between early stopping (false-alarm) and detection delay (time to detect abrupt change). It offers the key insight that the quickest way to detect a change, at any false-alarm rate, is by playing the `most informative action', of an `oracle' who a priori knows the post-change distribution. This suggests that to quickly identify changepoints, we must direct our effort towards rapidly identifying informative actions. On a technical level, we develop a change-of-measure argument for nonstationary, adaptive change detection, that allows for relating the divergences between random trajectories until stopping to the divergence of probability laws under each action for any two problem instances.

\noindent{\bf $\epsilon$-Greedy Change Detector (\egcd).} We propose \egcd, which, at a high level, uses a small amount of forced exploration to identify informative actions. The forced exploration allows for rapid convergence towards informative actions, and playing these actions minimizes detection delay. Our \egcd\, is based on the generalized max-likelihood/likelihood ratio principle which is utilized to estimate parametric changes. To prove detection delay bounds we draw upon key insights of \egcd. We first interpret the scheme in terms of competing parallel `queues', where each queue corresponds to a candidate post-change parameter, collects `arrivals' which are log-likelihood ratios of observations, and cannot go negative. The true parameter is the queue which enjoys the highest growth rate after a change, and the detection delay is the time required for it to dominate and become the `longest queue'. The dynamics of the queues can be related to nonstationary random walks with drifts. Using these insights, we prove that the expected detection delay of \egcd\, at low false alarm rates mirrors our information theoretic lower bounds thus establishing optimality of our method.

\noindent {\bf Experiments.} We perform numerical experiments of \egcd\, on synthetic and real datasets and show that under variations of changepoints, anomalies, and action sets, we realize gains due to adaptive sensing.


\section{Related Work} \label{sec:related}

Classical quickest change detection, dating back to the pioneering works of Page~\cite{page} and Lorden \cite{lorden1971procedures}, studies the problem of deciding when to stop (akin to declaring a change) while sequentially observing a data stream. In this context, the CUSUM algorithm \cite{page}, in which a change is announced as soon as a likelihood ratio statistic exceeds a threshold, has been shown to enjoy optimal performance. 

Our work is motivated by constraints on data collection in multi-stream time-series data, necessitating the design of an adaptive sampling rule in addition to the stopping rule. Similar to our work, Zhang and Mei  \cite{zhang2020bandit} also propose a sequential sampling  method for quickest change detection based on the well-known Shiryaev-Roberts-Pollak scheme (see, e.g., \cite{Basseville}). However, their theoretical analysis is rather limited in its scope. In particular, they argue ``in principle" that their method will control false alarms for sufficiently large thresholds, and will {\em eventually} (i.e., after infinite time) choose sensors where changes manifest, without any explicit characterization of detection delays. In contrast, we derive fundamental detection delay under a false alarm rate constraint, thus fully characterizing the BQCD problem. Additionally, their framework is not general enough to handle correlated or structured anomalies, which is a significant aspect of both our theoretical and experimental investigations.

In other related work, Gundersen et al  \cite{gundersen2021active} propose to extend the framework of Bayesian Online Change Detection (BOCD) \cite{adams2007bayesian} to incorporate costs in making decisions for real-time data acquisition in multi-fidelity sensing scenarios. Different from this perspective, our approach is frequentist, and does not impose action-specific costs and distributional constraints on the underlying latent parameters or on the changepoints. Furthermore, we derive finite-time performance bounds, which is not a focus of this work.

\section{Problem Setup} \label{sec:psetup}
{\bf General Setup.} We consider the following discrete-time model for the bandit quickest changepoint detection problem. At each decision round $t \in \mathbb{N} := \{1, 2, \ldots\}$, the parameter $\theta_t$ governs the distribution of the observation taken in that round. Let $\nu \in \mathbb{N}$ denote the round at which the change in behavior occurs, so that $\theta_t = \theta_0$ for $t < \nu$ and $\theta_t = \theta^* \in \ThetaOne$ for $t \geq \nu$. At each round $t$, a learner decides to either (1) stop and output that a change has taken place, denoted by the random variable $U_t = 1$, or (2) continue (denoted by $U_t = 0$) and play an action or sensing decision $A_t$ from an action set $\mathcal{A}$. Upon playing this action the learner obtains an observation $X_t$, sampled independent of the past, from a probability distribution $\Pobs{\theta_t}{\cdot}{A_t}$, which depends on both the current system parameter $\theta_t$ and the current action $A_t$. The stopping decision and sensing action ($U_t$ and $A_t$) are assumed to be chosen in a causal manner, i.e., depending on all past information $U_0, A_1, X_1, U_1, \ldots, A_{t-1}, X_{t-1}$, along with potentially independent internal randomness. The learner is aware, a priori, of the pre-change parameter $\theta_0$, the post-change parameter set $\ThetaOne$ and the observation distribution structure, i.e., $\left\{\Pobs{\theta}{\cdot}{a}\right\}_{\theta, a}$ with $a\in {\cal A}$ denotes fixed actions. 
Crucially, the change time $\nu$ and the specific post-change parameter $\theta^*$ are not known in advance and must be `learnt' in order to perform well.

The {stopping time} of the learner is defined to be $\tau = \min\{t \geq 0: U_t = 1\}$, and the main performance metric that we are interested in is the  detection delay, which is the random variable $(\tau - \nu)^+ = \max\{\tau - \nu, 0\}$. Specifically, we are interested in designing sampling and stopping rules so that (a) when no change occurs ($\nu = \infty$), $\tau$ should be at least a specified time delay with high probability, which is a measure of false alarm rate, and (b) when a change occurs ($\nu < \infty$), the expected detection delay should be small implying quick detection of change.

\noindent {\bf Illustrative Example.} For illustration, we elaborate on the scenario depicted in Fig.~\ref{fi:bqcd}. We consider $N$ `locations' as shown, and the observation at location $j$ at time $t$ follows:
$$Y_j(t)=\theta \mathbbm{1}_{[t \geq \nu] \wedge [j \in S]} + W_j(t),\,\,t\in \mathbb{Z}^+,\,\,j\in [N],$$ 
where $\left\{W_j(t)\right\}_{j,t}$ is a collection of standard normal random variables and $\mathbbm{1}_{[\cdot]}$ denotes the indicator function. This is a temporal extension of the model in \cite{qian14}. The model with the parameter $\theta \neq 0$ signifies an elevated mean after time $t \geq \nu \in \mathbb{Z}^+$ at locations $S \subseteq [N]$. The learner is agnostic to $\theta$, the time $\nu$ as well as which set $S$ among a family of known subsets ${\cal S} \subseteq \{0,1\}^N$ exhibits elevated activity. The learner at time $t$ can probe locations by means of sensing actions $A_t \in {\cal A} \subset \{0,1\}^N$, and observe a scalar measurement, $X_t \sim P_{\theta_t}(\cdot \mid A_t) \sim {\cal N}\left(\theta_t \frac{|A_t \cap S|}{\sqrt{|A_t|}},\,1 \right)$, with $\theta_t=0$ for $t < \nu$ and for $t \geq \nu$, $\theta_t =\theta \in \Theta^{(1)}=\mathbb{R}\setminus \{0\}$. The normalization factor $\sqrt{|A_t|}$ expresses the fact that the same SNR is maintained across different probes $A_t$ (see \cite{qian14}).

{\it General (Structured) Scenarios.} Depending on the application, locations maybe endowed with a neighborhood network structure, and change maybe observable in a $K$-neighborhood of the target location, and in such cases, ${\cal S}$ are constrained to be $K$-neighborly sets. Similarly, to allow for diverse probing modes, we could allow the action set to include actions that aggregate signal across multiple locations.

\noindent {\bf Notation.} For a post-change parameter $\theta \in \ThetaOne$, we use $\OptAction{\theta}$ to denote the most {\em informative} action for $\theta$ w.r.t.  $\theta^{(0)}$, i.e., $\OptAction{\theta} \in \argmax_{a \in \mathcal{A}} \kl{\Pobs{\theta}{\cdot}{a} }{\Pobs{\theta_0}{\cdot}{a}}$, where $\kl{\cdot}{\cdot}$ stands for the Kullback-Leibler divergence. To lighten notation, in the sequel we often abbreviate the distribution $ \Pobs{\theta}{\cdot}{a}$ to $\theta(a)$, and thus  $\kl{\Pobs{\theta}{\cdot}{a}}{\Pobs{\theta_0}{\cdot}{a}}$ to $\kl{\theta(a)}{\theta_0(a)}$. The simultaneous subscript-superscript notation $z_i^j$ is often used to represent the sequence $(z_i, z_{i+1}, \ldots, z_j)$. For any $\theta \in \ThetaOne$ and $n \in \mathbb{N}$, we use $\mathbb{P}^{(n, \theta)}$ to denote the probability measure on the process $U_0, A_1, X_1, U_1, A_2, X_2, \ldots$ induced by the learner's decisions, when the pre-change parameter $\theta_0$ changes to $\theta$ at time $n$. We also denote by $\mathbb{P}^{(\infty)}$ the probability measure as above under the no-change situation ($\nu=\infty$).

We define the filtration $({\cal F}_t)_{t\geq 0}$ to be the $\sigma$-algebra generated by all the past random variables up to time $t$, i.e., ${\cal F}_t=\sigma(\{(A_s,X_s), 0\leq s\leq t)\})$. An (admissible) change detector is an algorithm for which the stopping decision $U_t \in \{0, 1\}$ and action $A_t \in \cal{A}$ are ${\cal F}_{t-1}$-measurable at all rounds $t$.

{\bf Objective Specification.} Our BQCD framework involves a false alarm rate constraint specified by a tuple $(m,\alpha)$, where $m \in \mathbb{Z}^+$ denotes an a priori fixed time such that the probability of an admissible change detector stopping before time $m$ under the null hypothesis (no change) is at most $\alpha \in (0,1)$, namely, $\probSup{(\infty)}{\tau < m} \leq \alpha$. This kind of false alarm duration specification for the stopping time ($\geq m$ with probability $\geq 1-\alpha$) has been commonly employed in the literature on classical (non-adaptive) change detection (see e.g., \cite{lai1998information}), to ensure that the comparative performance rules out early triggering algorithms.
Let ${\cal D}_{m,\alpha}$ be the class of admissible change detectors with the $(m,\alpha)$ property stated above. As such, our goal is to find detectors in ${\cal D}_{m,\alpha}$ that minimize the detection delay when a change indeed occurs.

{\bf Informal Statement of Our Results.} Our main result is a sharp ``optimal characterization'' for expected detection delay in the limit of vanishing false alarms ($\alpha \rightarrow 0$, $m \to \infty$). The result, stated below, involves the new  $\epsilon$-Greedy-change-detector ($\epsilon$-GCD) algorithm in which, in each round $t$, we choose an action uniformly at random from the set ${\cal A}$ with probability $\epsilon$, or choose an action that is `most informative' for an estimated post-change parameter with probability $1-\epsilon$.  {\bf Objective Specification.} Our BQCD framework involves a false alarm rate constraint specified by a tuple $(m,\alpha)$, where $m \in \mathbb{Z}^+$ denotes an a priori fixed time such that the probability of an admissible change detector stopping before time $m$ under the null hypothesis (no change) is at most $\alpha \in (0,1)$, namely, $\probSup{(\infty)}{\tau < m} \leq \alpha$. This kind of


\begin{theorem}[informal]
Consider the limiting situation as $\alpha \rightarrow 0$ and $m = \omega\left(\log \frac{1}{\alpha} \right) \to \infty$. Then, any admissible change detector in ${\cal D}_{m,\alpha}$ must suffer $$\liminf_{\alpha \to 0} \frac{\mathbb{E}^{(\nu,\theta^*)}[(\tau-\nu)^+]}{  \log\frac{1}{\alpha}} \geq \frac{1}{\max_{a} D({\theta^*(a)}||{\theta^{(0)}(a)})}.$$ On the other hand, the $\epsilon$-GCD algorithm, suitably tuned to be in ${\cal D}_{m,\alpha}$, achieves   $$\limsup_{\alpha \to 0} \frac{\mathbb{E}^{(\nu,\theta^*)}[(\tau-\nu)^+]}{  \log\frac{1}{\alpha}} \leq \frac{8}{(1-\epsilon)}  \frac{1}{\max_{a} D({\theta^*(a)}||{\theta^{(0)}(a)})  }.$$ 

\end{theorem}



\section{Fundamental lower bounds on bandit change detection performance}
\label{sec:lowerbound}

Before embarking upon the design of bandit changepoint algorithms, it is worth understanding what the limits of bandit change detection performance are, due to the stochastic and partial nature of the problem's information structure. We expose in this section a universal lower bound on the detection delay of {\em any} bandit changepoint algorithm with a false alarm rate above a given level. 


\begin{theorem}
	\label{thm:lowerbound}
	Let $0 < \alpha \leq \frac{
	1}{10}$ and $m \geq 1$. 
	For any bandit changepoint algorithm satisfying  $\probSup{(\infty)}{\tau < m} \leq \alpha$, we have $\mathbb{E}^{(1, \theta^*)}\left[\tau  \right] \geq \min \left\{  \frac{\frac{1}{20} \log \frac{1}{\alpha}}{\max_{a \in \mathcal{A}}  \kl{\theta^*(a)}{\theta^{(0)}(a)}}, \frac{m}{2}
	\right\} \quad \forall \theta^* \in \ThetaOne$.

\end{theorem}
The result implies that in the `small' false alarm rate ($\alpha$) regime with `large' $m = \omega(\log (1/\alpha))$, e.g., $m = \log^2 (1/\alpha)$, we have that any algorithm meeting the false alarm rate property $\probSup{(\infty)}{\tau \geq m} \geq 1-\alpha$ must suffer a detection delay at least $\Omega\left(  \frac{\log \frac{1}{\alpha}}{\max_{a \in \mathcal{A}}  \kl{\theta^*(a)}{\theta^{(0)}(a)}} \right)$ 
when a change occurs at time $1$ (a similar, `anytime' lower bound holds for detection delay at $\nu \in \mathbb{N}$, see appendix). 

{\bf False alarm-detection Delay Tradeoff.} Theorem \ref{thm:lowerbound} shows that a basic tradeoff exists between the false alarm rate or early stopping rate in case of no change, on one hand, and the detection delay after a true change occurs, on the other. Specifically, it is impossible to stop `too early', i.e., before time  $\Omega\left(  \frac{\log \frac{1}{\alpha}}{\max_{a \in \mathcal{A}}  \kl{\theta^*(a)}{\theta^{(0)}(a)}} \right)$, after a true change if one wishes to stop `too late' under no change, i.e., $\probSup{(\infty)}{\tau \geq m} \geq 1-\alpha$. We also note that when there is no adaptive sampling of actions (i.e., $|\mathcal{A}| = 1 $), then the lower bound reduces to the form of a a standard lower bound for the classical changepoint detection problem \cite{lai1998information}.

{\bf Information Structure.} This is perhaps the most valuable implication of the lower bound. The quantity $\kl{\theta^*(a)}{\theta^{(0)}(a)}$, in the denominator, can be interpreted as the amount of information that playing an action $a$ provides (on average) in order to detect a change from $\theta_0$ to $\theta^*$. Consequently, the quickest way to detect a change is by playing the `most informative action', $\argmax_{a \in \mathcal{A}}  \kl{\theta^*(a)}{\theta^{(0)}(a)}$. This can be viewed as the sampling strategy of an `oracle' who knows the post-change parameter $\theta^*$ in advance. Unfortunately, this information is not known a priori for a causal algorithm. However, we will see that this can in fact be learnt along the trajectory and the lower bound can be attained order-wise by a suitable learning algorithm that we propose. 




\noindent {\bf Proof Sketch for Theorem \ref{thm:lowerbound}.} The main idea is a measure change argument, adapted to our non-stationary change point setting, from literature on sample complexity bounds for bandit best arm identification  \cite{garivier-kaufmann16optimal-best-arm-fixed-confidence}. On one hand, if the algorithm in consideration is  very `lazy' to begin with, i.e., $\probSup{(1, \theta^*)}{\tau \geq m}$ is at least a constant, say $1/2$, then we immediately get $\expectSup{(1, \theta^*)}{\tau} \geq \frac{1}{2} \cdot m$. On the other hand, if the algorithm is `active', i.e., $\probSup{(1, \theta^*)}{\tau < m} \geq 1/2$, then the KL divergence between the laws of the indicator random variable $\indic{\tau < m}$ in the instances $(1, \theta^*)$ and $(\infty, \cdot)$ is `large' (at least a constant times $\log(1/\alpha)$) owing to the hypothesis that $\probInf{\tau < m}$ is `small' (at most $\alpha$). But by the data processing inequality, this divergence is bounded above by the divergence between the distributions of the entire trajectory $U_0, A_1, X_1, U_1, A_2, X_2, \ldots$ under the two instances, which can be seen to be equivalent to the quantity $\sum_{a \in \mathcal{A}} \kl{\theta^*(a)}{\theta(a)} \expectSup{(1, \theta^*)}{N_{\tau}(a)}$, and which is further bounded above by $\expectSup{(1, \theta^*)}{\tau} \max_{a \in \mathcal{A}} \kl{\theta^*(a)}{\theta(a)}$. See appendix for a complete proof. 


	


\section{\egcd\, Algorithm}
\label{sec:algorithmdesc}


We describe the \egcd\, adaptive sensing algorithm for the bandit quickest change detection problem. 
%
%
The lower bound in Section \ref{sec:lowerbound} suggests that it is beneficial to infer the target post-change parameter, so that playing the most informative action for it can yield the best possible detection delay performance. This is the key principle underlying the design of the \egcd\, algorithm (Algorithm \ref{algo:epsgreedy}). 

At a high level, \egcd\,  uses a small amount of forced exploration along with `greedy' exploitation to play sensing actions. Specifically, it computes, at each round $t$, a maximum likelihood estimate (MLE) of the post-change parameter based on the generalized likelihood ratio test (GLRT) principle. It then plays either a randomly chosen action, if the current slot is an exploration slot, or a `greedy', i.e., most informative, action for the estimated post-change parameter, if it is an exploitation slot.

The MLE of the post-change parameter, in round $t$, admits an interpretation as the longest `queue' corresponding to some parameter in $\ThetaOne$. To see this, notice that the MLE for the pair $(\nu, \theta^*)$, given all previous data in exploration rounds can be written as $\argmax_{\theta \in \Theta^{(1)}, 1 \leq v \leq t} \prod_{\ell = 1}^{v-1} f_{\theta_0}(X_\ell|A_\ell)^{E_\ell}  \prod_{\ell=v}^{t-1} f_{\theta}(X_\ell|A_\ell)^{E_\ell} = \argmax_{\theta \in \Theta^{(1)}, 1 \leq v \leq t} \sum_{\ell=v}^{t-1} E_\ell \log \frac{f_{\theta}(X_\ell|A_\ell)}{f_{\theta_0}(X_\ell|A_\ell)}$, %
with $f_{\theta}(\cdot|a)$ taken to be the probability density or mass function of the distribution  $\mathbb{P}_{\theta}(\cdot|A_\ell)$ (assuming one exists). Observe now that for each candidate post-change parameter $\theta$,  the inner maximum over $v$,  $\ExploreQueue_t(\theta) := \argmax_{1 \leq v \leq t} \sum_{\ell=v}^{t-1} E_\ell \log \frac{f_{\theta}(X_\ell|A_\ell)}{f_{\theta_0}(X_\ell|A_\ell)}$, evolves as a `queue'\footnote{This is also known as the Lindley recursion equation in queueing theory.} with arriving work $E_\ell \log \frac{f_{\theta}(X_\ell|A_\ell)}{f_{\theta_0}(X_\ell|A_\ell)}$ at time slot $\ell$; in other words, $\ExploreQueue_{t+1}(\theta) = \left( \ExploreQueue_t(\theta) + E_t \log \frac{f_{\theta}(X_t|A_t)}{f_{\theta_0}(X_t|A_t)} \right)^+$.

We define the algorithm using general `processing'  functions $g_\theta$ in place of the log-likelihood ratios $\log \frac{f_\theta}{f_{\theta_0}}$ above. Broadly speaking, the functions $g$ should ideally be chosen with the hope that (a) $g_\theta(X_\ell, A_\ell)$ is negative in expectation before the change time ($\ell < \nu$), and (b) $g_\theta(X_\ell, A_\ell)$ is large and positive in expectation for $\theta = \theta^*$, the true change parameter, after the change ($\ell \geq \nu$).

The stopping rule that \egcd\, uses is based on the generalized likelihood ratio (GLR)-type statistic. 
It is the largest of an ensemble of evolving exploitation-data queues $\ExploitQueue_t(\theta)$, which mirrors the definition of $\ExploreQueue_t(\theta)$ but with $1-E_t$ instead of $E_t$.

\subsection{Theoretical guarantees for the \egcd\, algorithm}
\label{sec:theoryresults}
In this section, we present and discuss theoretical guarantees on the false alarm rate and detection delay performance of the \egcd\, algorithm of Section \ref{sec:algorithmdesc}.

\begin{wrapfigure}[30]{R}{0.5\textwidth}
\vspace{-32pt}
\begin{minipage}{0.5\textwidth}
\begin{algorithm}[H]
	\renewcommand\thealgorithm{1}
	\caption{\egcd}\label{algo:epsgreedy}
	\begin{algorithmic}[1]
		\STATE \textbf{Input:}  $\epsilon \in [0,1]$,  $\theta_0$,  $\Theta^{(1)}$, $\beta \geq 1$,  $\pi$, Observation function $g_\theta(x, a)$ $\forall (\theta, x, a) \in \ThetaOne \times \mathcal{X} \times \mathcal{A}$.
		
		\STATE {\bf Init:} $\ExploreQueue_1(\theta) \leftarrow 0$, $\ExploitQueue_1(\theta) \leftarrow 0$ $\forall \theta \in \ThetaOne$ \hfill \COMMENT{CUSUM statistics for exploration and exploitation per $\theta$}
		
		\FOR{round $t = 1, 2, 3, \ldots$}
		\IF{$\max_{\theta \in \ThetaOne} \ExploitQueue_t(\theta) \geq \log \beta$}
		\STATE {\bf break} \hfill \COMMENT{Stop sampling and exit}
		\ENDIF
		\STATE Sample $E_t \sim \text{Ber}(\epsilon)$ independently
		\IF{$E_t == 1$} 
		\STATE Play action $A_t \sim \pi$ independently  \hfill \COMMENT{Explore}
		\STATE Get observation $X_t$
		\STATE Set $\forall \theta \in \ThetaOne:$ $\ExploreQueue_{t+1}(\theta) \leftarrow \left( \ExploreQueue_t(\theta) + g_\theta(X_t, A_t) \right)^+$ \hfill \COMMENT{Update exploration CUSUM statistics}
		\ELSE
		\STATE Compute $\hat{\theta}_t = \argmax_{\theta \in \Theta^{(1)}} \ExploreQueue_t(\theta)    $ \hfill\COMMENT{Most likely post-change distribution based on exploration data}
		\STATE Play action $A_t = \argmax_{a \in \cal{A}} \kl{\hat{\theta}_t(a)}{\theta^{(0)}(a)}$  
		\STATE Get observation $X_t$
		\STATE Set $\forall \theta \in \ThetaOne:$ $\ExploitQueue_{t+1}(\theta) \leftarrow \left( \ExploitQueue_t(\theta) + g_\theta(X_t, A_t) \right)^+$ \hfill \COMMENT{Update exploitation CUSUM statistics}
		\ENDIF
		
		\ENDFOR
	\end{algorithmic}
\end{algorithm}
\end{minipage}
\end{wrapfigure}

We make the following assumptions on the parameter space and observation distributions in order to derive performance bounds for the algorithm. 


\begin{assumption}
	\label{ass:finite}
	The post-change parameter set $\Theta^{(1)}$ is finite, i.e., $|\Theta^{(1)}| < \infty$. 
\end{assumption}
This assumption is made primarily for ease of analysis, whereas the algorithm is defined even for arbitrary parameter spaces. Specifically, it allows for easy control of the fluctuations of an ensemble of (drifting) random walks via a union bound over the parameter set. While we believe that this can be relaxed to handle general parameter spaces via appropriate netting or chaining arguments, this chiefly technical task is left to future investigation.

\begin{assumption}
	\label{ass:discrimination}
	Every post-change parameter is detectable by some action, i.e.,  $\forall \theta \in \ThetaOne \,  \exists a \in \mathcal{A}: \kl{\theta(a)}{\theta^{(0)}(a)} > 0$. Moreover, any two post-change parameters are separable by some action, i.e., $\forall \theta, \theta' \in \ThetaOne \,  \exists a \in \mathcal{A}: \kl{\theta(a)}{\theta'(a)} > 0$.
\end{assumption}
This is a basic identifiability requirement of the setting, without which some parameter changes could be completely undetectable. Put differently, one can only hope to detect changes that the sensing set can tease apart.



\begin{assumption}[Bounded marginal KL divergences]
\label{ass:boundedKL}
There is a constant $D_{\max}$ satisfying $\kl{\theta_1(a)}{\theta_2(a)} \leq D_{\max}$ for each $a \in \mathcal{A}$, $\theta_1, \theta_2, \in \ThetaOne$.
\end{assumption}

This assumption is easily met, for instance, if all log likelihood ratios are bounded, or if the observations are modelled as Gaussians. 
In this case log likelihood ratios are linear functions of the observation.

\begin{assumption}
\label{ass:LLR}
    Every observation probability distribution  $\Pobs{\theta}{\cdot}{a}$, for $a \in \mathcal{A}$ and $\theta \in \ThetaOne \cup \{\theta_0\}$, has either a  density\footnote{with respect to a standard reference measure, e.g., Lebesgue measure} or a mass function, denoted by $f_{\theta}(\cdot | a)$. Moreover, there exists $r > 0$ such that for any  $\theta, \theta', \theta'' \in \ThetaOne \cup \{\theta_0\}$, and $a \in \mathcal{A}$, the log likelihood ratio $\frac{f_{\theta'}(X | a)}{f_{\theta''}(X | a)}$ is  $r$-subgaussian \footnote{A random variable $X$ is $r$-subgaussian under $X \sim \mathbb{P}$ if $\forall \lambda \in \mathbb{R}$: $\mathbb{E}[e^{\lambda ( X -\mathbb{E}[X] )  }] \leq e^{\lambda^2 r/2}$.} under $X \sim \Pobs{\theta}{\cdot}{a}$.
\end{assumption}
This assumption is common in the statistical inference literature; we use it to be able to control the fluctuations of the exploration and exploitation CUSUM statistics in the algorithm via standard subgaussian concentration tools. 
A concrete example of a setting that meets Assumptions \ref{ass:finite}-\ref{ass:LLR} is the linear observation model described in Section~\ref{sec:psetup} and Fig.~\ref{fi:bqcd}.
We are now in a position to state our key theoretical result.

\begin{theorem}[False alarm and detection delay for general change point]
	\label{thm:epsgreedygenchange}
	Under assumptions  \ref{ass:finite}-\ref{ass:LLR}, the following conclusions hold for \egcd\, (Algorithm \ref{algo:epsgreedy}) run with the log-likelihood  observation function  $g_\theta(x,a) \equiv \log \frac{f_{\theta}(x|a)}{f_{\theta_0}(x|a)}$. 
	\begin{enumerate}[wide,nosep]
		\item (Time to false alarm) Let $\alpha \in (0,1)$ and $m \in \mathbb{N}$. If the stopping threshold $\beta$ is set as $\beta \geq \frac{m \SizeThetaOne}{\alpha}$, then the stopping time satisfies  $\probInf{\tau < m} \leq \alpha$.
		
		\item (Detection delay) 
		For a change from $\theta_0$ to $\theta^* \in \ThetaOne$ occurring at time $\nu \in \mathbb{N}$,
		\begin{align} \expectSup{(\nu,\theta^*)}{(\tau - \nu)^+} &\leq 
		\frac{8 \log \beta}{ (1-\epsilon)\max_{a} \kl{\theta^*(a)}{\theta^{(0)}(a)}  } + O\left( \text{poly}\left(\frac{1}{\epsilon} \right), \expectInf{\ExploreQueue_\nu}, \mathcal{P} \right) , \label{eq:detdelaybound}
		\end{align}
		provided $\pi(a^*_{\theta^*}) > 0$, with $\ExploreQueue_\nu := \left( \ExploreQueue_\nu(\theta) \right)_{\theta \in \ThetaOne}$ denoting the explore queue statistics for all parameters and $\mathcal{P} \equiv \left( \Pobs{\theta}{\cdot}{a} \right)_{\theta \in \{\theta_0\} \cup \ThetaOne}$  all the  observation distributions. 
	\end{enumerate}
\end{theorem}

{\it Remark.} We note that $\beta$ is an internal, algorithmic parameter, whose setting is spelt out in order that the $(m,\alpha)$ false alarm constraint can be met. Thus, the first part of Theorem 3 is a ‘correctness’ statement\footnote{This is, at a high level, akin to specifying a PAC algorithm’s correctness requirement and measuring its sample complexity on the other hand, i.e., insist that its returned answer be $\epsilon$-correct with probability at least $1-\delta$, and then analyze how long it takes to learn and return an answer. The former corresponds to the false alarm constraint and the latter to measuring detection delay here.}

{\bf Interpretation of the Detection Delay Bound.} The first term in the detection delay bound \eqref{eq:detdelaybound} can be interpreted as the time for an {\em oracle} fixed-action strategy (e.g., CUSUM), that always plays the most informative action for $\theta^*$, to stop. The second term, on the other hand, is a bound on the time taken by the algorithm to {\em learn} to play the most informative action for $\theta^*$ (details follow in proof sketch). We omit the precise dependence of the second term on the problem structure here for brevity, but detail it explicitly in the appendix. Specifically, for small forced exploration rates $\epsilon$, the second term depends on the information geometry of the problem approximately as $\frac{1}{\epsilon^4} \sum_{\theta \in \ThetaOne: a^*_{\theta} \neq a^*_{\theta^*}}   \frac{1}{\Delta_\theta^4 }$. Here, for each candidate post-change parameter $\theta \in \ThetaOne$, its `gap' $\Delta_\theta$ is a measure of how difficult it is for the algorithm to eliminate $\theta$ as an estimate for the true post-change parameter $\theta^*$ during this parameter-learning phase. We formally define it as $\Delta_\theta := \klavg{\theta^*}{\theta_0} -  \frac{1}{2}(\klavg{\theta^*}{\theta_0} + \left( \klavg{\theta^*}{\theta_0} - \klavg{\theta^*}{\theta} \right)^+)$, where $\klavg{\theta^*}{\theta} := \sum_{a \in \mathcal{A}} \pi(a) \kl{\theta^*(a)}{\theta(a)}$ is the average information divergence between parameters $\theta^*, \theta$ offered by playing from the exploration distribution $\pi$. With finer analysis, the dependence on gaps and $\epsilon$ could be improved. We omit it for simplicity. 

We now turn to the (additive \& linear; see the appendix) dependence on $\expectInf{\ExploreQueue_\nu}$ of the second term. The insight as to why this arises is as follows. The `learning' phase for $\theta^*$, after a change takes place at time $\nu$, can be seen as a `race' between competing log-likelihood queues of all the parameters vying to become the MLE $\hat{\theta}_t$. At the change time, each non-ground truth queue $\ExploreQueue_\nu(\theta)$ is, on average, at level $\expectInf{\ExploreQueue_\nu(\theta)}$, representing its initial `advantage' over the ground-truth queue $ \ExploreQueue_\nu(\theta^*)$ which in the worst case could be at level $0$. Thus, the difference $\expectInf{\ExploreQueue_\nu(\theta)} - \ExploreQueue_\nu(\theta^*) \leq \ExploreQueue_\nu(\theta)$ is the extra amount of `time work' that the $\theta^*$ queue must do to overcome the $\theta$ queue. A special case of the result is for $\nu = 1$ in which case all queues start out at $0$. For general $\nu$, one can establish that $\expectInf{\ExploreQueue_\nu(\theta)}$ must be finite, by noticing that each queue is non-negative and accumulates arrivals with negative mean before the change, and applying standard queueing stability arguments. This also indicates that the detection delay bound is roughly invariant given the location of the change point $\nu$. 

{\bf Optimality at low false-alarms.} The result implies that in the `small' time to false alarm regime, i.e., $\alpha \to 0$, $m = \omega(\log(1/\alpha))$ and   $\log(m) = o(\log(\alpha))$, the detection delay when $\beta$ is set as above is dominated by the first term: $\expectSup{(\nu,\theta^*)}{\tau} = O\left( \frac{\log \beta}{(1-\epsilon)\max_{a} \kl{\theta^*(a)}{\theta^{(0)}(a)}} \right)$. 
%
%
This matches, order-wise, the universal lower bound of Theorem \ref{thm:lowerbound} up to the algorithm-dependent multiplicative factor $1/(1-\epsilon)$, which may be interpreted as a `penalty' for the forced exploration mechanism.

{\bf Benefits of Adaptivity.} Let us consider a concrete  setting to elucidate the detection delay bound. In this example, there are $d$ sensing actions, represented by the canonical basis vectors in $\mathbb{R}^d$. The pre-change parameter is $0 \in \mathbb{R}^d$, and there are $d$ candidate post-change parameters, each of which is a canonical basis vector $e_i$, $i \in [d]$, multiplied by a constant $\delta$. The observation from applying action $a$ at system parameter $\theta$ is Gaussian distributed with mean $\ip{a}{\theta}$ and variance $1$. Thus, the aim is to detect a (sparse) change of magnitude $\delta$ in one of the $d$ coordinates in $\theta_0 = 0$, when at any time only a single coordinate can be noisily sensed. Suppose $\theta^* = \delta e_1$ without loss of generality, for which the most informative action (in fact, the only informative one) is $a^*_{\theta^*} = e_1$. The first term in the detection delay given by Theorem \ref{thm:epsgreedygenchange} is then $O\left(\frac{\log (1/\alpha)}{(1-\epsilon) \delta^2 } \right)$. This is a factor of $d$ smaller than $\Omega\left(d \cdot  \frac{\log (1/\alpha)}{(1-\epsilon) \delta^2 } \right)$ that a standard CUSUM rule with non-adaptive uniform sampling over coordinates would achieve -- this is seen by applying the lower bound (Theorem \ref{thm:lowerbound}) to the case of a single (trivial action) where the divergence in the denominator reduces to the {\em average} divergence across all coordinates: $O(\delta^2/d)$. For estimating the second term, we calculate that for each $\theta = \delta e_i$ with $i \neq 1$, $\Delta_\theta = \klavg{\theta^*}{\theta_0} -  \frac{(\klavg{\theta^*}{\theta_0} + \left( \klavg{\theta^*}{\theta_0} - \klavg{\theta^*}{\theta} \right)^+)}{2} = \frac{\delta^2}{2d} - \frac{\frac{\delta^2}{2d} - \left(\frac{\delta^2}{2d} - \frac{\delta^2}{d} \right)^+  }{2} = \frac{\delta^2}{2d} - \frac{\frac{\delta^2}{2d} - 0  }{2}= \frac{\delta^2}{4d}$. Thus, the second term representing the time to learn the optimal sensing action scales as $O\left( \frac{(d-1)d^4}{\epsilon^4 \delta^8} \right) = O\left( \frac{d^5}{\epsilon^4 \delta^8} \right)$. Although we have not optimized dependence on $d, \epsilon$ and $\delta$, the overall bound still gives an idea of how the active sensing problem gets harder as the dimension $d$ grows or as the minimum change amount $\delta$ changes, making it akin to finding a `needle in a haystack'.

{\bf Sketch of the Proof of Theorem \ref{thm:epsgreedygenchange}.} This section lays down the key arguments involved in the proof of Theorem \ref{thm:epsgreedygenchange} for bounding the false alarm time and detection delay of the \egcd\, algorithm.

{\bf 1. Bounding the probability of early stopping under no change.} 
The stopping time $\tau$ for \egcd, by line \ref{line:stopping} in Algorithm \ref{algo:epsgreedy}, is equivalent to the first instant $t$ when the  worst-case (over $\ThetaOne$) CUSUM statistic computed over exploitation data, i.e., $\max_{\theta \in \ThetaOne} \ExploitQueue_t(\theta) \equiv \max_{\theta \in \ThetaOne} \max_{1 \leq s \leq t} \prod_{\ell = s}^{t-1} \left( \frac{f_{\theta}(X_\ell|A_\ell)}{f_{\theta_0} (X_\ell|A_\ell)} \right)^{(1-E_\ell)}$ 
exceeds the level $\beta \geq 1$. Under no change, each observation $X_\ell$ is distributed as $\Pobs{\theta_0}{\cdot}{A_\ell}$ given $A_\ell$, and the product $\prod_{\ell = s}^{t-1} \left( \frac{f_{\theta}(X_\ell|A_\ell)}{f_{\theta_0} (X_\ell|A_\ell)} \right)^{(1-E_\ell)}$, over $t = s, s+1, s+2, \ldots$, behaves as a (standard) likelihood ratio martingale with unit expectation under the appropriate filtration. Hence, the chance that the largest among this ensemble of martingales, one for each $\theta \in \ThetaOne$ and starting time $s \in [m]$, rises above $\beta$ before time $m$ is bounded using a union bound and Doob's inequality for each individual martingale, yielding the probability bound $\frac{m \SizeThetaOne}{\beta}$. 

{\bf 2. Control of the detection delay.} Suppose that the change takes place at time $\nu = 1$ to the post-change parameter $\theta^* \in \ThetaOne$. The proof strategy is to show, with high probability, that $\tau$ is no more than $t_1 + t_2$, where
\begin{itemize}[wide, nosep]\vspace{-5pt}
    \item $t_1$ is an upper bound for the time taken for the plug-in estimate $\hat{\theta}_t$, of the post-change parameter, to `settle' to $\theta^*$. In other words, we show that after time $t_1$ it is very unlikely that an action other than $a^*_{\theta^*} = \argmax_{a \in \cal{A}} \kl{\theta^{(0)}(a)}{\theta^*(a)}$ is played in every exploitation round.
    
    \item $t_2$ is an upper bound for the time taken for the worst-case CUSUM statistic $\max_{\theta \in \ThetaOne} \ExploitQueue_t(\theta)$ to grow to the level $\beta$ {\em assuming that the optimal (i.e., most informative) action $a^*_{\theta^*}$  is always played at all exploitation rounds}. 
\end{itemize}

For clarity of exposition, we will assume that we are in the setting of {\em linear measurements} in $\mathbb{R}^d$ with {\em additive standard Gaussian noise} -- this makes KL divergences easy to interpret as Euclidean distances -- and that the changepoint is at $\nu = 1$.

\noindent {\bf Step 1: Finding $t_1$ (time to learn $\theta^*$).} We observe that the MLE $\hat{\theta}_t$ at time $t$ can be written as the parameter $\theta$ associated with the largest stochastic process $\max_{v=1}^{t-1} \log J_{v,t-1}(\theta)$, one for each $\theta$: $\hat{\theta}_t = \argmax_{\theta \in \Theta^{(1)}} \max_{v=1}^{t-1} \log J_{v,t-1}(\theta)$, where $J_{v,t-1}(\theta) := \prod_{\ell = v}^{t-1} \left( \frac{f_{\theta}(X_\ell|A_\ell)}{f_{\theta_0}(X_\ell|A_\ell)} \right)^{E_\ell}$.

Under the post-change distribution $\mathbb{P}^{(1, \theta^*)}$, $\log J_{v,t-1}(\theta)$ can be seen to evolve (as a function of $t$) as a random walk with drift  $\epsilon (\norm{\theta^*}_H^2 - \norm{\theta^* - \theta}_H^2)$. Here,  $\norm{\cdot}_H$ is the usual matrix-weighted Euclidean norm in $\mathbb{R}^d$, governed by how much different directions are explored in expectation\footnote{This matrix also arises commonly in linear design of experiments as the Kiefer-Wolfowitz matrix \cite{lattimore2020bandit}.} by the exploration distribution $\pi$: $H = \sum_{a \in \mathcal{A}} \pi(a) a a^T$. Note that $\norm{\theta^* - \theta}_H^2$ here is the KL divergence  between the distributions of the observation that results when an action sampled from $\pi$ is played, under parameters $\theta^*$ and $\theta$.

The preceding discussion implies that the drift is the largest (and positive) for the random walk corresponding to the true post-change parameter $\theta = \theta^*$. The remainder of this part of the proof uses martingale concentration tools (subgaussian Chernoff and maximal Hoeffding bounds) to find the time $t_1$ at which the fastest-growing random walk, $\log J_{v,t}(\theta^*)$ dominates all other `competing' random walks $\log J_{v,t}(\theta)$, $\theta \neq \theta^*$. 

\noindent {\bf Step 2: Finding $t_2$ (time to stop under optimal action plays).} In a manner similar to that of Step 1, we observe that the logarithm of the log-likelihood ratio for $\theta^*$ w.r.t. $\theta_0$ computed over only {\em exploitation} rounds, i.e., $\log \prod_{\ell = v}^{t-1} \left( \frac{f_{\theta}(X_\ell|A_\ell)}{f_{\theta_0}(X_\ell|A_\ell)} \right)^{1-E_\ell}$, evolves as a random walk with drift rate $(1-\epsilon)\norm{\theta^*}_H $. A Chernoff bound can then be used to control the probability of the `bad' event that this growing random walk has {\em not} crossed the level $\log \beta$ in a certain time duration $t_2$ ($t_2$ does not appear explicitly in the main derivation).


\section{Experiments}
\label{sec:expts}

\begin{wraptable}[12]{r}{8cm}
\vspace{-20pt}
\begin{tabular}{ccccc}\\\toprule  
  Size    & Oracle & \egcd+ &  \egcd & URS \\ \midrule
 10  & $30\pm 5 $ & $98\pm 62$ & $112\pm 74$ & $306\pm 76$ \\  \midrule
 15   & $31\pm5 $ & $129\pm 95$ & $158\pm 119 $ & $ 460\pm 115 $\\  \midrule
 20    &$31\pm 5 $ & $163\pm128$ & $196\pm156 $ & $616 \pm153 $\\  \midrule
 25   & $31\pm 5 $ & $191\pm154$ & $253\pm216 $ & $764 \pm194 $\\  \midrule
\end{tabular}
\caption{\footnotesize Observed mean and standard deviation for the simulated stopped time for varying graph-sizes with pointy action set ${\cal A}_1$, for change occurring at $\nu=40$.}
\label{tab.size}
\end{wraptable}

Our goal in this section is to illustrate various aspects of our theory through experiments on synthetic environments, and to explore the performance of the proposed \egcd\, algorithm on a setting based on a real-world dataset. All experiments were performed on a laptop with Intel Core i5 CPU and 8GB of RAM, and take under an hour to execute. The closely related prior works on this topic ~\cite{zhang2020bandit,gundersen2021active} are more specific and are not compatible with many of our structured scenarios (see Sec.~\ref{sec:related} and Appendix~\ref{app:relatedwork}). As alternatives, we propose several natural baselines and compare against our method to benchmark performance.

{\bf Uniform Random Sampling (URS).} At every round we sample each action in ${\cal A}$ uniformly at random. Our goal with URS is to calibrate adaptation gains.\\
{\bf $\epsilon$-GCD (Ours).} This is our proposed scheme. A point to be noted here (as seen from Algorithm~\ref{algo:epsgreedy} is that the data used for making a stopping decision ("exploitation") and that for estimating the changed distribution ("exploration") are non-overlapping, and as such leads to inefficient data usage.\\
{\bf $\epsilon$-GCD+.} To reduce the aforementioned data inefficiency, we consider the variant, the parameter estimation part also utilizes both data collected during exploration rounds and the data collected for the exploitation rounds.\\
{\bf Oracle.} Here we consider an omniscient strategy in that the algorithm is aware of both the pre-change and post-change distributions ahead of time, and consequently does not require exploration to estimate the post-change parameter. On the other hand the method has no knowledge of stopping time, and thus the problem reduces to the conventional quickest change detection. The Oracle characterizes the gold-standard in that no adaptation is required, and it gives the absolute smallest detection delay for any BQCD problem.

{\bf Synthetic Experiments.} 
We conduct synthetic experiments for the model introduced in Section~\ref{sec:psetup} under the bold item titled ``illustrative example.''  
We explore how average detection delay varies under controlled variation of various parameters such as changepoint location, dimensionality and structure of the ambient space, and the type of action sets. Our objective is twofold: (a) Illustrate gains due to adaptivity of proposed \egcd\, over the non-adaptive method, where actions are chosen uniformly at random; (b) Demonstrate ``near'' optimality by baselining against Oracle.




We report results for the case when the ambient space is a line graph (see Illustrative example in Sec.~\ref{sec:psetup}). Nodes are interpreted as physical locations, and take values in $[N]$. Nodes $j,k$ are connected if $|j-k|=1$. Each node $n \in [N]$ offers a Gaussian-distributed observation depending on the changepoint $\nu \in \mathbb{N}$. 



{\it Isolated and Structured Anomalies.} The vector of change parameters $\theta=[\theta_n]$ are of two types: (a) Isolated singleton change, namely, $\theta_n \in \{0,\,1\}$ and $\sum_{n \in [N]}\theta_n=1$; (b) Structural changes, i.e., $\mbox{Supp}(\theta)=\{n \in [N]: \theta_n \neq 0 \}$ is k-connected set with $\theta \in \{0,1\}^N$.

{\it Diffuse and `Pointy' Action sets.} In a parallel fashion we consider two types of action sets. The set ${\cal A}_1$,  is pointy, namely, $|\mbox{Supp}({\cal A}_1)|=1$, which allows probing only single nodes. The set ${\cal A}_2$ is diffuse where only a connected subset of nodes can be queried. In either case, the observation received on an action is as described in Sec.~\ref{sec:psetup}. 

\begin{floatingfigure}[r]{0.5\textwidth}
\centering
\includegraphics[width=0.4\textwidth]{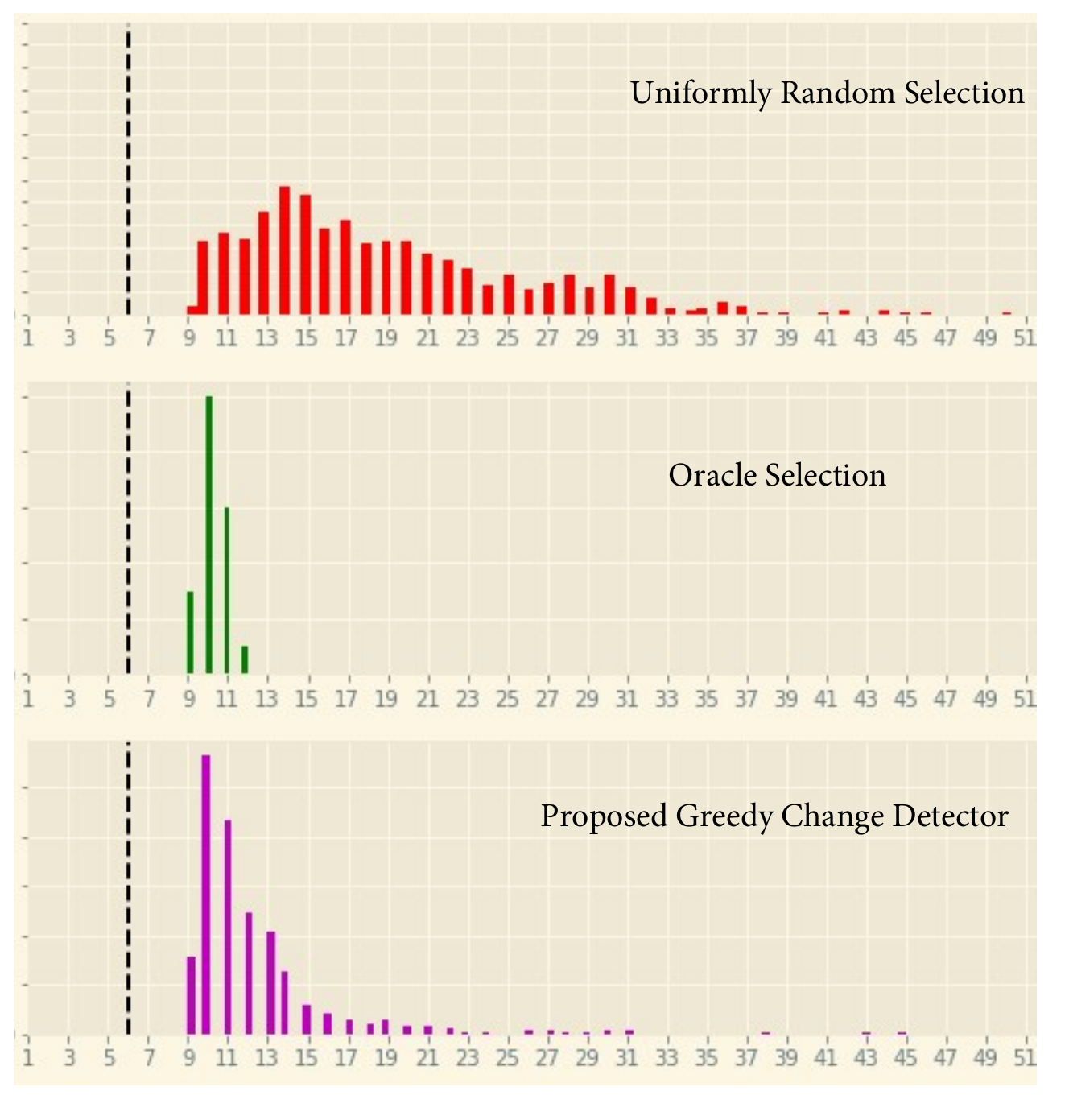}
\caption{\footnotesize Audio based change detection of machine anomaly: Histogram of stopping times by URS, Oracle, and \egcd. Histograms for Oracle and \egcd\, demonstrates clear adaptivity gains.}
\label{fig.audio_hist}

\end{floatingfigure}

{\bf Key Findings.}
We report results for pointy actions and isolated anomalies, and defer other settings to the appendix. We choose $\sigma^2=0.5$, $\beta$ such that the false alarms are about $1\%$ for the Oracle, and a forced exploration rate $\epsilon=0.2$. Our results are averaged over \textcolor{black}{5000} Monte-Carlo runs. With this choice for $\beta$ we did not observe false alarms in any of the algorithms.\\
{\it Gains from Adaptivity.}
The fact that we perform substantially better than URS uniformly across all of the experiments demonstrates our gains through adaptive sampling.\\
{\it Data Inefficiency.}
The fact that the proposed method closely tracks \egcd+ suggests that our decoupled scheme is capable of fully leveraging the observations, and the cost of decoupled exploration is small. \\
{\it Optimality.}
The fact that the proposed method tracks Oracle qualitatively across all experiment setups indicates that our method is close to optimal. This is because through adaptivity we achieve performance similar to a competitor who requires no adaptation (recall that oracle knows the change parameters and thus does not need to explore). In point of fact we note in passing that any other method would at best lie between our proposed method and the Oracle. \\
{\it Variation with Changepoint $\nu$.} For a fixed graph size, we observed that the average expected delay is relatively constant for all methods, which is consistent with Theorem~\ref{thm:epsgreedygenchange}.

{\bf Audio based recognition of machine anomalies.}
We experiment using the MIMII audio dataset \cite{purohit_harsh_2019_3384388}. 
 Detailed specifics are in the appendix. The dataset has four machines (ID00,ID02,ID04,ID06), each equipped with audio sensors recording the health of the machine. There are three types of anomaly (rail damage, loose belt, no grease)  which can occur at any time, in any machine. Corresponding to each anomaly there is an audio stream, and the anomaly occurs in one of the four machines at an arbitrary time. For each machine, the dataset contains audio-streams of 1000 normal and 300 abnormal files, and each audio-stream is about 10 seconds long. 

{\it Audio Processing.} For each audio-stream we train autoencoders on normal data using mel-spectrogram features,  and fit Gaussians to the reconstruction errors. This results in pre- and post-change parameters for each machine's autoencoded reconstruction error score, corresponding to normal and abnormal operation. 


{\it Experiment.} To simulate BQCD, we introduced anomalies in machine bearing ID00 as follows. We concatenated 6 normal files and 54 abnormal files chosen uniformly at random from machine ID00. For the other machines we concatenated 60 normal files at random. The 60 files correspond to 600 seconds. Our changepoint corresponds to 6th file, which we denote as $\nu=6$ and our task is to detect this change. Note that the machine ID and the changepoint are both not known to the learner. Our results are depicted as histograms for changepoints of anomaly detection in Fig.~\ref{fig.audio_hist}. As observed, we notice that \egcd's performance is close to Oracle both in mean and distribution, while URS exhibits larger delay and significant variance. Appendix \ref{app:audioexpts} presents histograms for a larger changepoint.

\newpage
\begin{ack}
A. G. and B. L. were supported by the Aerospace Network Research Consortium (ANRC) Grant on Airplane IOT Data Management. 
V. S. was supported by the Army Research Office Grant W911NF2110246, and National Science Foundation grants CCF-2007350 and CCF-1955981. The authors are grateful to Rajesh Sundaresan, Himanshu Tyagi  and Manjunath Krishnapur for helpful discussions, and to the anonymous program committee members for their valuable feedback. 

\end{ack}



\printbibliography 

\section*{Checklist}


\begin{enumerate}

\item For all authors...
\begin{enumerate}
  \item Do the main claims made in the abstract and introduction accurately reflect the paper's contributions and scope?
    \answerYes{See Section \ref{sec:intro}.}
  \item Did you describe the limitations of your work?
    \answerYes{See Section \ref{sec:theoryresults} for assumptions under which the results hold.} 
  \item Did you discuss any potential negative societal impacts of your work?
    \answerNo{As a primarily theoretical paper concerned with efficiency, our work does not have any meaningful negative impacts on society beyond the larger issues of how it is employed.}
  \item Have you read the ethics review guidelines and ensured that your paper conforms to them?
    \answerYes{}
\end{enumerate}

\item If you are including theoretical results...
\begin{enumerate}
  \item Did you state the full set of assumptions of all theoretical results?
    \answerYes{See Section \ref{sec:theoryresults}.}
	\item Did you include complete proofs of all theoretical results?
    \answerYes{See the supplementary material for all proofs.}
\end{enumerate}

\item If you ran experiments...
\begin{enumerate}
  \item Did you include the code, data, and instructions needed to reproduce the main experimental results (either in the supplemental material or as a URL)?
    \answerYes{We have included all the code and instructions to run it as part of the supplementary material. The dataset (MIMII audio sounds) used is large, and is available publicly.}
  \item Did you specify all the training details (e.g., data splits, hyperparameters, how they were chosen)?
    \answerYes{We have specified the details of the experiments and parameter choices in the supplementary material.}
	\item Did you report error bars (e.g., with respect to the random seed after running experiments multiple times)?
    \answerYes{We report either standard deviations or empirical distributions in our experiments.}
	\item Did you include the total amount of compute and the type of resources used (e.g., type of GPUs, internal cluster, or cloud provider)?
    \answerYes{See Section \ref{sec:expts}.}
\end{enumerate}

\item If you are using existing assets (e.g., code, data, models) or curating/releasing new assets...
\begin{enumerate}
  \item If your work uses existing assets, did you cite the creators?
    \answerYes{See subsection on ``Audio based recognition of machine anomalies" in Section \ref{sec:expts}.}
  \item Did you mention the license of the assets?
    \answerYes{See the citation to the MIMII audio dataset in Section \ref{sec:expts}.}
  \item Did you include any new assets either in the supplemental material or as a URL?
    \answerNA{No new assets were created.}
  \item Did you discuss whether and how consent was obtained from people whose data you're using/curating?
    \answerNA{We have not used data from people.}
  \item Did you discuss whether the data you are using/curating contains personally identifiable information or offensive content?
    \answerNA{The only measured data used in the paper are those of machine sounds.}
\end{enumerate}

\item If you used crowdsourcing or conducted research with human subjects...
\begin{enumerate}
  \item Did you include the full text of instructions given to participants and screenshots, if applicable?
    \answerNA{}
  \item Did you describe any potential participant risks, with links to Institutional Review Board (IRB) approvals, if applicable?
    \answerNA{}
  \item Did you include the estimated hourly wage paid to participants and the total amount spent on participant compensation?
    \answerNA{}
\end{enumerate}

\end{enumerate}



\clearpage
\appendix
{\bf {\Large Appendix}} \\
\section{Proof of Theorem \ref{thm:lowerbound}}
We in fact establish the following more general  result.

\begin{theorem}
	\label{thm:lowerboundanytime}
	 Let $0 < \alpha \leq \frac{
	1}{10}$, $m \geq 1$ and $\nu \in \mathbb{N}$. 
	If a bandit changepoint algorithm satisfies  $\probSup{(\infty)}{\tau < \nu + m \given \nu \leq \tau} \leq \alpha$, then for any $\theta^* \in \ThetaOne$, 
	\[ \expectSup{(\nu, \theta^*)}{\tau - \nu \given \tau \geq \nu} \geq \min \left\{ \frac{\frac{1}{20} \log \frac{1}{\alpha}}{ \max_{a \in \mathcal{A}}  \kl{\theta^*(a)}{\theta_0(a)} }, \frac{m}{2} \right\}. \]
\end{theorem}

Note that Theorem \ref{thm:lowerbound} is  the special case\footnote{We assume that $\tau \geq 1$ with probability $1$.} of $\nu = 1$.

We first recall and/or put down some preliminaries before embarking upon the proof. 

\noindent {\bf Definition: Problem instance.} For any changepoint time $\nu \in \{1, 2, \ldots\} \cup \{\infty\}$ and post-change parameter $\theta \in \Theta^{(1)}$, we call the pair $(\nu, \theta)$ an {\em instance} of the bandit changepoint detection problem. Note that if $\nu = \infty$, then it is immaterial what the post-change parameter $\theta$ is, since there is effectively no change in the distribution; thus we will omit $\theta$, or write $(\infty, *)$, if $\nu = \infty$ for ease of notation. The instance, along with the sampling algorithm and (known) $\theta^{(0)}$, completely determines the distribution of trajectories generated by the operation of the algorithm. \\

\noindent {\bf Bandit changepoint detection algorithm:} Recall the definition of an algorithm for the bandit changepoint detection problem: It is a rule that maps the history $I_t$  of actions and observations to (1) a decision $U_t \in \{0,1\}$ to stop playing actions, and (2) if not stopping ($U_t = 0$), then an action $A_t$ to play in round $t$. Here, 
\[I_t = (V_0, U_0, A_1, X_1, V_1, U_1, A_2, X_2, V_2, \ldots, V_{t-2}, U_{t-2}, A_{t-1}, X_{t-1}, V_t), \]
where at any time instant $s \geq 0$, $V_s$ represents independent, internal randomness available to the algorithm, $U_s$ is an indicator random variable for the event that the algorithm decides to stop playing before taking the $(s+1)$-st action (i.e., after playing $s$ actions), and $X_s$ is the observation from playing arm $A_s$ in round $s$.

\begin{proof}[Proof of Theorem \ref{thm:lowerboundanytime}]

We first establish an auxiliary lemma about the explicit form for the divergence of the conditional distribution of a trajectory.

Let $\text{Law}^{(i, \theta)}_E(I_\tau)$ denote the probability distribution of the random trajectory $I_\tau$ conditioned on the event $E$, when the algorithm is run on the instance $(i, \theta)$. 

\begin{lemma}
	\label{lem:divergencecond}
	For any parameter $\theta \in \Theta^{(1)}$, $\eta \in \mathbb{N}$ and $E \in \sigma(I_{\eta})$,
	\begin{equation*}
		\kl{\text{Law}^{(\eta, \theta)}_E(I_\tau)}{\text{Law}^{(\infty)}_E(I_\tau)} = \sum_{a \in \mathcal{A}} \kl{\theta(a)}{\theta_0(a)} \expectSup{(\eta, \theta)}{ \left( N_\tau(a) - N_{\eta-1}(a) \right)^+ \given E  }  . 
	\end{equation*}
\end{lemma}

\begin{proof}
For the sake of convenience we show the argument assuming that the trajectory $I_\tau$ is a discrete random object, i.e., it has a probability mass function. (It is straightforward, but notationally heavier, to extend it to the case of general measures by using Radon-Nikodym derivatives.) 

We have
\begin{align*}
    &\kl{\text{Law}^{(\eta, \theta)}_E(I_\tau)}{\text{Law}^{(\infty)}_E(I_\tau)} \\
    &= \sum_{\omega \equiv (v_0, u_0, a_1, x_1, v_1, \ldots, v_{\tau - 1}, u_\tau = 1) \in E} \probSup{(\eta, \theta)}{I_\tau = \omega \given E} \log \frac{\probSup{(\eta, \theta)}{I_\tau = \omega \given E}}{\probInf{I_\tau = \omega | E}} \\
    &\stackrel{(a)}{=} \sum_{\omega \equiv (v_0, u_0, a_1, x_1, v_1, \ldots, v_{\tau - 1}, u_\tau = 1) \in E} \probSup{(\eta, \theta)}{I_\tau = \omega \given E} \log \frac{\probSup{(\eta, \theta)}{I_\tau = \omega}}{\probInf{I_\tau = \omega }} \\
    &= \sum_{\omega \in E}  \probSup{(\eta, \theta)}{I_\tau = \omega \given E } \, \log \frac{\probSup{(\eta, \theta)}{v_0}\probSup{(\eta, \theta)}{u_0 | v_0} \probSup{(\eta, \theta)}{a_1 | u_0, v_0} \probSup{(\eta, \theta)}{x_1 | a_1} \cdots}{\probInf{v_0}\probInf{u_0 | v_0} \probInf{a_1 | u_0, v_0} \probInf{x_1 | a_1} \cdots} \\
    &\stackrel{(b)}{=} \sum_{\omega \in E} \probSup{(\eta, \theta)}{I_\tau = \omega \given E} \log \frac{\probSup{(\eta, \theta)}{x_1 | a_1} \probSup{(\eta, \theta)}{x_2 | a_2} \cdots}{ \probInf{x_1 | a_1} \probInf{x_2 | a_2} \cdots} \\
    &= \sum_{\omega \in E} \probSup{(\eta, \theta)}{I_\tau = \omega \given E} 
   \sum_{t=1}^{\tau-1} \log \frac{\probSup{(\eta, \theta)}{x_t(\omega) | a_t(\omega)}}{ \probInf{x_t(\omega) | a_t(\omega)}} \\
   &\stackrel{(c)}{=} \frac{1}{\probInf{E}} \expectSup{(\eta, \theta)}{\sum_{t=1}^{\tau-1} \log \frac{\probSup{(\eta, \theta)}{X_t | A_t}}{ \probInf{X_t | A_t}} \indic{E} }. 
\end{align*}
Here, $(a)$ and $(c)$ both follow because $\probSup{(\eta, \theta)}{E} = \probInf{E}$ due to $E \in \sigma(I_\eta)$, and $(b)$ is because the algorithm's decisions and internal randomness $v_0, u_0, a_1, a_2$, etc. do not depend on the probability distribution of the environment generating the observations. We continue further, writing
\begin{align*}
	&\probInf{E} \cdot \kl{\text{Law}^{(\eta, \theta)}_E(I_\tau)}{\text{Law}^{(\infty)}_E(I_\tau)} \\
	&= \expectSup{(\eta, \theta)}{\indic{E} \sum_{t=1}^{(\eta-1) \wedge (\tau-1)} \log \frac{\probSup{(\eta, \theta)}{X_t | A_t}}{ \probInf{X_t | A_t}} + \indic{E} \sum_{t=(\eta-1) \wedge (\tau-1) + 1}^{\tau-1} \log \frac{\probSup{(\eta, \theta)}{X_t | A_t}}{ \probInf{X_t | A_t}} } \\
	&= \expectSup{(\eta, \theta)}{\indic{E}  \sum_{t=\eta \wedge \tau}^{\tau-1} \log \frac{\probSup{(\nu, \theta)}{X_t | A_t}}{ \probInf{X_t | A_t}} \sum_{a \in \mathcal{A}} \indic{A_t=a} }, 
\end{align*}
because $E \in \sigma(I_\eta)$. Thus,	
\begin{align*}
	&\probInf{E} \cdot \kl{\text{Law}^{(\eta, \theta)}_E(I_\tau)}{\text{Law}^{(\infty)}_E(I_\tau)} \\
    &= \sum_{a \in \mathcal{A}} \sum_{t=\eta \wedge \tau}^{\tau-1} \expectSup{(\eta, \theta)}{ \log \frac{\probSup{(\nu, \theta)}{X_t | A_t}}{ \probInf{X_t | A_t}}  \indic{A_t=a, E} } \\
    &= \sum_{a \in \mathcal{A}} \sum_{t=\eta \wedge \tau}^{\tau-1} \expectSup{(\eta, \theta)}{ \indic{A_t=a, E} \expectSup{(\eta, \theta)}{   \log \frac{\probSup{(\eta, \theta)}{X_t | A_t}}{ \probInf{X_t | A_t}} \bigg\vert \indic{A_t=a, E} } } \\
    &\stackrel{(d)}{=} \sum_{a \in \mathcal{A}} \sum_{t=\eta \wedge \tau}^{\tau-1} \expectSup{(\eta, \theta)}{ \indic{A_t=a, E} \kl{\theta(a)}{\theta_0(a)} } \\
    &= \sum_{a \in \mathcal{A}} \kl{\theta(a)}{\theta_0(a)} \expectSup{(\eta, \theta)}{ \indic{E} \left( N_\tau(a) - N_{\eta-1}(a) \right)^+  },
\end{align*}
where $(d)$ is due to $E \in \sigma(I_\eta)$ and the Markov property of the algorithm's trajectory. This completes the proof. 
\end{proof}

Returning to the proof of the theorem, we split the analysis into two cases depending on the value of the conditional probability $\probSup{(\nu, \theta^*)}{\tau < \nu + m \given \nu \leq \tau }$ of stopping before an additional time $m$ after having crossed the actual changepoint $\nu$.
	
	\noindent {\bf Case 1:} $\probSup{(\nu, \theta^*)}{\tau < \nu + m \given \nu \leq \tau } \geq \frac{1}{2}$. In this case, applying the data processing inequality for KL divergence \cite{cover1999elements} to the two (conditional) input distributions $\text{Law}^{(\nu, \theta^*)}_{\nu \leq \tau}(I_\tau) \equiv \probSup{(\nu, \theta^*)}{I_\tau \in \cdot \given \nu \leq \tau}$ and $\text{Law}^{(\infty)}_{\nu \leq \tau}(I_\tau) \equiv \probInf{I_\tau \in \cdot \given \nu \leq \tau}$, with the data processing function $f(I_\tau) := \indic{\nu \leq \tau < \nu + m}$, yields
	\begin{align*}
		&\kl{\text{Law}^{(\nu, \theta^*)}_{\nu \leq \tau}(I_\tau)}{\text{Law}^{(\infty)}_{\nu \leq \tau}(I_\tau)} \\
		&\geq \kl{ \text{Ber}\left( \probSup{(\nu, \theta^*)}{\tau < \nu + m \given \nu \leq \tau} \right) }{ \text{Ber} \left( \probSup{(\infty)}{\tau < \nu + m \given \nu \leq \tau}  \right) }.
	\end{align*}
	Together with Lemma \ref{lem:divergencecond} for the event $E := \{ \nu \leq \tau \}$, this gives
	\begin{align*}
		&\sum_{a \in \mathcal{A}} \kl{\theta^*(a)}{\theta_0(a)} \expectSup{(\nu, \theta^*)}{  \left( N_\tau(a) - N_{\eta-1}(a) \right)^+ \given \nu \leq \tau } \\
		&\geq \kl{ \text{Ber}\left( \probSup{(\nu, \theta^*)}{\tau < \nu + m \given \nu \leq \tau} \right) }{ \text{Ber} \left( \probSup{(\infty)}{\tau < \nu + m \given \nu \leq \tau}  \right) } \\
		&\stackrel{(a)}{\geq} \mathbb{P}^{(\nu, \theta^*)}[ \tau < \nu + m \given \nu \leq \tau] \log \frac{\mathbb{P}^{(\nu, \theta^*)}[\tau < \nu + m \given \nu \leq \tau]}{\mathbb{P}^{(\infty)}[ \tau < \nu + m \given \nu \leq \tau]} - \ln 2 \\
		&\stackrel{(b)}{\geq} \frac{1}{2} \log \frac{1/2}{\alpha} - \ln 2 \geq \frac{1}{20} \log\frac{1}{\alpha},
	\end{align*}
	by Lemma \ref{lem:logineq}, where $(a)$ is due to \cite{garivier2019explore} and $(b)$ is by hypothesis. We now divide both sides by $\sum_{a \in \mathcal{A}}\expectSup{(\nu, \theta^*)}{  \left( N_\tau(a) - N_{\eta-1}(a) \right)^+ \given \nu \leq \tau  } = \expectSup{(\nu, \theta^*)}{  \left( \tau - \nu \right)^+ \given \nu \leq \tau }$, and use the fact that the maximum is at least a convex combination, to get 
	\begin{align*}
		\max_{a \in \mathcal{A}} \kl{\theta^*(a)}{\theta_0(a)} &\geq \frac{1}{20} \frac{\log (1/\alpha)}{ \expectSup{(\nu, \theta^*)}{  \left( \tau - \nu \right)^+ \given \nu \leq \tau } },
	\end{align*}
	giving one part of the theorem.
	
	\noindent {\bf Case 2:} $\probSup{(\nu, \theta^*)}{\tau < \nu + m \given \nu \leq \tau } < \frac{1}{2}$. In this case, we have 
	\begin{align*}
		\expectSup{(\nu, \theta^*)}{  \left( \tau - \nu \right)^+ \given \nu \leq \tau } &\geq m \cdot \probSup{(\nu, \theta^*)}{\tau \geq \nu + m \given \nu \leq \tau } \geq \frac{m}{2},
	\end{align*}
	giving the other part of the theorem.

\begin{lemma}
	\label{lem:logineq}
	For $0 < x < \frac{
	1}{10}$, we have $\frac{1}{2} \log\frac{1}{2x} - \log 2 \geq \frac{1}{20} \log \frac{1}{x}$.
\end{lemma}
\begin{proof}
	The proof is by basic calculus. 
\end{proof}

\end{proof}

\section{Proof of Theorem \ref{thm:epsgreedygenchange}}

\subsection{Time to false alarm}

The \egcd\, algorithm stops at the first time $t$ when the largest exponentiated `exploitation queue' CUSUM statistic, i.e., 
\[ \max_{\theta \in \ThetaOne} \exp \left( \ExploitQueue_t(\theta) \right) := \max_{\theta \in \ThetaOne} \max_{1 \leq s \leq t} \prod_{\ell = s}^{t-1} \left( \frac{f_{\theta}(X_\ell|A_\ell)}{f_{\theta_0} (X_\ell|A_\ell)} \right)^{(1-E_\ell)},  \]
exceeds $\beta$. 

For $m \in \mathbb{N}$, let us compute
\begin{align}
	&\probInf{\tau < m} = \probInf{\exists t < m: \max_{\theta \in \ThetaOne} \max_{1 \leq s \leq t} \prod_{\ell = s}^{t-1} \left( \frac{f_{\theta}(X_\ell|A_\ell)}{f_{\theta_0} (X_\ell|A_\ell)} \right)^{(1-E_\ell)} \geq \beta} \nonumber \\
	&\leq \probInf{\exists \theta \in \ThetaOne, s \in [m]: \max_{s \leq t \leq m} \prod_{\ell = s}^{t-1} \left( \frac{f_{\theta}(X_\ell|A_\ell)}{f_{\theta_0} (X_\ell|A_\ell)} \right)^{(1-E_\ell)} \geq \beta} \nonumber \\
	&\leq \sum_{\theta \in \ThetaOne} \sum_{s=1}^m \probInf{ \max_{s \leq t \leq m} \prod_{\ell = s}^{t-1} \left( \frac{f_{\theta}(X_\ell|A_\ell)}{f_{\theta_0} (X_\ell|A_\ell)} \right)^{(1-E_\ell)} \geq \beta }, \label{eq:stopInf}
\end{align}
by a union bound. 

\begin{lemma}
\label{lem:LLRmartingale}
For each fixed $s \in [m]$ and $\theta \in \ThetaOne$, the likelihood ratio process $\left\{ \prod_{\ell = s}^{t-1} \left( \frac{f_{\theta}(X_\ell|A_\ell)}{f_{\theta_0} (X_\ell|A_\ell)} \right)^{(1-E_\ell)}  \right\}_{t \geq s}$ is a mean-$1$ martingale under the measure $\mathbb{P}^{(\infty)}$ and with respect to the filtration $(\mathcal{F}_t)_{t \geq s}$, where 
\[\forall t \geq s \quad \mathcal{F}_t := \sigma(E_s, A_s, X_s, E_{s+1}, A_{s+1}, X_{s+1}, \ldots, E_{t-1}, A_{t-1}, X_{t-1}). \]
\end{lemma}

\begin{proof}
Taking the conditional expectation of the $t$-th term of the process w.r.t. $\mathcal{F}_{t-1}$, we get 
\begin{align*}
    &\expectInf{ \prod_{\ell = s}^{t-1} \left( \frac{f_{\theta}(X_\ell|A_\ell)}{f_{\theta_0} (X_\ell|A_\ell)} \right)^{(1-E_\ell)} \bigg\vert \mathcal{F}_{t-1} } \\
    &= \prod_{\ell = s}^{t-2} \left( \frac{f_{\theta}(X_\ell|A_\ell)}{f_{\theta_0} (X_\ell|A_\ell)} \right)^{(1-E_\ell)} \expectInf{ \left( \frac{f_{\theta}(X_{t-1}|A_{t-1})}{f_{\theta_0} (X_{t-1}|A_{t-1})} \right)^{(1-E_{t-1})} \bigg\vert \mathcal{F}_{t-1} }.
\end{align*}
The conditional expectation on the right hand side satisfies
\begin{align*}
    &\expectInf{ \left( \frac{f_{\theta}(X_{t-1}|A_{t-1})}{f_{\theta_0} (X_{t-1}|A_{t-1})} \right)^{(1-E_{t-1})} \bigg\vert \mathcal{F}_{t-1} } \\
    &= \probInf{E_{t-1} = 1} + \probInf{E_{t-1} = 0}\cdot \expectInf{  \frac{f_{\theta}(X_{t-1}|A_{t-1})}{f_{\theta_0} (X_{t-1}|A_{t-1})}  \bigg\vert \mathcal{F}_{t-1} },
\end{align*}
where we have used the independence of the exploration decision $E_{t-1}$ from the past. By iterated expectation, we have
\begin{align*}
    \expectInf{  \frac{f_{\theta}(X_{t-1}|A_{t-1})}{f_{\theta_0} (X_{t-1}|A_{t-1})}  \bigg\vert \mathcal{F}_{t-1} } &= \expectInf{ \expectInf{  \frac{f_{\theta}(X_{t-1}|A_{t-1})}{f_{\theta_0} (X_{t-1}|A_{t-1})} \bigg\vert A_{t-1}, \mathcal{F}_{t-1}  }  \bigg\vert \mathcal{F}_{t-1} } \\
    &= \expectInf{ 1   \given \mathcal{F}_{t-1} },
\end{align*}
establishing the result.
\end{proof}
Thanks to Lemma \ref{lem:LLRmartingale}, we can apply Doob's maximal inequality to the non-negative likelihood ratio martingale above to get
\begin{align*}
    \probInf{ \max_{s \leq t \leq m} \prod_{\ell = s}^{t-1} \left( \frac{f_{\theta}(X_\ell|A_\ell)}{f_{\theta_0} (X_\ell|A_\ell)} \right)^{(1-E_\ell)} \geq \beta } &\leq \frac{1}{\beta}.
\end{align*}
Together with \eqref{eq:stopInf}, this implies
\begin{align*}
	\probInf{\tau < m} &\leq \sum_{\theta \in \ThetaOne} \sum_{s=1}^m \probInf{ \max_{s \leq t \leq m} \prod_{\ell = s}^{t-1} \left( \frac{f_{\theta}(X_\ell|A_\ell)}{f_{\theta_0} (X_\ell|A_\ell)} \right)^{(1-E_\ell)} \geq \beta } \leq \frac{\SizeThetaOne m}{\beta}.
\end{align*}
Therefore, using a stopping threshold satisfying $\beta \geq \frac{m \SizeThetaOne }{\alpha}$ guarantees the false alarm property $\probInf{\tau < m} \leq \alpha$.

\subsection{Detection delay}
\label{app:detectiondelay}

{\bf Preliminaries.} Let the true post-change parameter starting from an arbitrary time $\nu \in \mathbb{N}$ be equal to $\theta^* \in \ThetaOne$. 

Recall that the \egcd\, algorithm (Algorithm \ref{algo:epsgreedy}) makes, at round $t \geq 1$, the generalized maximum likelihood estimate
\begin{align*}
	\hat{\theta}_t &= \argmax_{\theta \in \Theta^{(1)}} \max_{v=1}^{t} \prod_{\ell = 1}^{v-1} f_{\theta_0}(X_\ell|A_\ell)^{E_\ell}  \prod_{\ell=v}^{t-1} f_{\theta}(X_\ell|A_\ell)^{E_\ell} \\
	&= \argmax_{\theta \in \Theta^{(1)}} \max_{v=1}^{t} \prod_{\ell = 1}^{v-1} f_{\theta_0}(X_\ell|A_\ell)^{E_\ell}  \prod_{\ell=v}^{t-1} f_{\theta}(X_\ell|A_\ell)^{E_\ell} \left/ \prod_{\ell = 1}^{t-1} f_{\theta_0}(X_\ell|A_\ell)^{E_\ell} \right. \\
	&= \argmax_{\theta \in \Theta^{(1)}} \max_{v=1}^{t} \prod_{\ell = v}^{t-1} \left( \frac{f_{\theta}(X_\ell|A_\ell)}{f_{\theta_0}(X_\ell|A_\ell)} \right)^{E_\ell} = \argmax_{\theta \in \Theta^{(1)}} \max_{v=1}^{t} J_{v,t-1}(\theta) = \argmax_{\theta \in \Theta^{(1)}} \max_{v=1}^{t} \log J_{v,t-1}(\theta),
\end{align*}
where we have denoted $J_{v,t-1}(\theta) := \prod_{\ell = v}^{t-1} \left( \frac{f_{\theta}(X_\ell|A_\ell)}{f_{\theta_0}(X_\ell|A_\ell)} \right)^{E_\ell}$ $\Leftrightarrow$ $\log J_{v,t-1}(\theta) = \sum_{\ell = v}^{t-1} E_\ell \log \frac{f_{\theta}(X_\ell|A_\ell)}{f_{\theta_0}(X_\ell|A_\ell)} = \sum_{\ell = v}^{t-1} U_\ell(\theta)$, where we have defined $U_\ell(\theta) \equiv U_\ell(\theta, \theta_0, X_\ell, A_\ell) := E_\ell \log \frac{f_{\theta}(X_\ell|A_\ell)}{f_{\theta_0}(X_\ell|A_\ell)}$. Also recall that $\ExploreQueue_t(\theta) = \max_{v=1}^{t} \log J_{v,t-1}(\theta) \geq 0$ represents the CUSUM-style  `exploration queue' statistic for each candidate parameter $\theta \in \ThetaOne$. 

{\bf Step 1: Bounding the time for the `right
 CUSUM queue $\ExploreQueue_t(\theta^*)$ to outstrip other  queues.}
 
Suppose $\theta \in \ThetaOne$ satisfies  $a^*_{\theta} \neq a^*_{\theta^*}$. For an arbitrary time $t \geq \nu$, we can write  
\begin{align}
    &\probSup{(\nu, \theta^*)}{\max_{i=1}^{t} \log J_{i,t-1}(\theta) \geq \max_{i=1}^{t} \log J_{i,t-1}(\theta^*) } = \probSup{(\nu, \theta^*)}{\max_{i=1}^{t} \sum_{\ell = i}^{t-1} U_\ell(\theta) \geq \max_{i=1}^{t} \sum_{\ell = i}^{t-1} U_\ell(\theta^*) } \nonumber \\
    &\leq \probSup{(\nu, \theta^*)}{ \max_{i=1}^{\nu} \sum_{\ell = i}^{\nu-1} U_\ell(\theta) + \max_{i=\nu}^{t} \sum_{\ell = i}^{t-1} U_\ell(\theta) \geq \max_{i=1}^{t} \sum_{\ell = i}^{t-1} U_\ell(\theta^*) } \quad \text{(by Lemma \ref{lem:partialsums})} \nonumber \\
    &\leq \probSup{(\nu, \theta^*)}{ \max_{i=1}^{\nu} \sum_{\ell = i}^{\nu-1} U_\ell(\theta) + \max_{i=\nu}^{t} \sum_{\ell = i}^{t-1} U_\ell(\theta) \geq  \sum_{\ell = \nu}^{t-1} U_\ell(\theta^*) } \quad \text{(since $\nu \in [t]$)}  \nonumber \\
    &= \probSup{(\nu, \theta^*)}{ \ExploreQueue_\nu(\theta) + \max_{i=\nu}^{t} \sum_{\ell = i}^{t-1} U_\ell(\theta) \geq  \sum_{\ell = \nu}^{t-1} U_\ell(\theta^*) }   \nonumber \\
    &= \expectSup{(\nu, \theta^*)}{ \probSup{(\nu, \theta^*)}{ \ExploreQueue_\nu(\theta) + \max_{i=\nu}^{t} \sum_{\ell = i}^{t-1} U_\ell(\theta) \geq  \sum_{\ell = \nu}^{t-1} U_\ell(\theta^*) \given \ExploreQueue_\nu(\theta) } }. \nonumber \\
    &= \expectInf{ \probSup{(\nu, \theta^*)}{ \ExploreQueue_\nu(\theta) + \max_{i=\nu}^{t} \sum_{\ell = i}^{t-1} U_\ell(\theta) \geq  \sum_{\ell = \nu}^{t-1} U_\ell(\theta^*) \given \ExploreQueue_\nu(\theta) } }. \label{eq:expectProb} 
\end{align}
The final equality is due to the fact that the inner conditional probability is a function of only $\ExploreQueue_\nu(\theta)$, whose distribution is identical under $\mathbb{P}^{(\nu, \theta^*)}$ and $\mathbb{P}^{(\infty)}$ because it is determined by actions and observations before the change time $\nu$.

We now make the crucial observation that for any $q \geq 0$,
\begin{align} 
&\probSup{(\nu, \theta^*)}{ Q_\nu(\theta) + \max_{i=\nu}^{t} \sum_{\ell = i}^{t-1} U_\ell(\theta) \geq  \sum_{\ell = \nu}^{t-1} U_\ell(\theta^*) \given Q_\nu(\theta) = q } \nonumber \\
&=\probSup{(\nu, \theta^*)}{ q + \max_{i=\nu}^{t} \sum_{\ell = i}^{t-1} U_\ell(\theta) \geq  \sum_{\ell = \nu}^{t-1} U_\ell(\theta^*) \given Q_\nu(\theta) = q } \nonumber \\
&= \probSup{(1, \theta^*)}{ q + \max_{i=1}^{t-\nu+1} \sum_{\ell = i}^{t-\nu} U_\ell(\theta) \geq  \sum_{\ell = 1}^{t-\nu} U_\ell(\theta^*)  }. \label{eq:Qnu}
\end{align}

The first equality above is  by simply substituting for $Q_\nu(\theta)$, but the second equality holds because (a) the random variables $E_\ell, X_\ell, A_\ell$ for $\ell \geq \nu$ are independent of $Q_\nu(\theta)$, by virtue of the independent forced exploration enforced in the algorithm, and (b) the probability distribution of exploration actions and their corresponding observations from round $\nu$ onward under $\mathbb{P}^{(\nu, \theta^*)}$ is the same as that of the observations and actions from round $1$ onward under $\mathbb{P}^{(1, \theta^*)}$. In other words, {\em we have rewound the time clock so that the change point is at time $1$ instead of time $\nu \geq 1$.} 

Going forward, to lighten notation, we use $\mathbb{E}$ and $\mathbb{P}$ instead of $\mathbb{E}^{(1,\theta^*)}$ and $\mathbb{P}^{(1,\theta^*)}$ in our calculations. We start by bounding the expectation on the right hand side of \eqref{eq:Qnu}: \begin{align}
    &\prob{ q + \max_{i=1}^{t-\nu+1} \sum_{\ell = i}^{t-\nu} U_\ell(\theta) \geq  \sum_{\ell = 1}^{t-\nu} U_\ell(\theta^*)   } \nonumber \\
    &\leq \prob{ \psi \geq  \sum_{\ell = 1}^{t-\nu} U_\ell(\theta^*) } + \prob{ q + \max_{i=1}^{t-\nu+1} \sum_{\ell = i}^{t-\nu} U_\ell(\theta) \geq  \psi }, \label{eq:rwprobscond}
\end{align}
where $\psi$ is chosen as follows. We first introduce the shorthand $\klshort{\theta_1}{\theta_2}{a} := \kl{\theta_1(a)}{\theta_2(a)}$. We then define, for each $\theta \in \ThetaOne \cup \{\theta_0\}$, 
\[ \klavg{\theta^*}{\theta} := \expectSup{(1, \theta^*)}{ \klshort{\theta^*}{\theta}{A_\ell} \given E_\ell = 1  } = \sum_{a \in \mathcal{A}} \pi(a) \klshort{\theta^*}{\theta}{a}   \]
to be the average KL divergence between $\theta^*$ and $\theta$, when an action is randomly sampled from the exploration distribution $\pi$. Intuitively, $\epsilon (\klavg{\theta^*}{\theta_0} - \klavg{\theta^*}{\theta})$ is the average rate of drift of the queue $\ExploreQueue_t(\theta)$ at any time after the changepoint $t \geq \nu$ within the exploration rounds; thus, the queue $\ExploreQueue_t(\theta^*)$ enjoys the highest possible drift rate upward.

Finally, we let 
\[ \psi := \epsilon (t-\nu) \cdot \frac{\klavg{\theta^*}{\theta_0} + \left( \klavg{\theta^*}{\theta_0} - \klavg{\theta^*}{\theta} \right)^+}{2}.\] 

We will also find it useful to define the `gap' of a parameter $\theta$ w.r.t. the true post-change parameter $\theta^*$ as  \[\Delta_\theta := \klavg{\theta^*}{\theta_0} -  \frac{\klavg{\theta^*}{\theta_0} + \left( \klavg{\theta^*}{\theta_0} - \klavg{\theta^*}{\theta} \right)^+}{2} = \frac{1}{2} \min \left( \klavg{\theta^*}{\theta_0}, \klavg{\theta^*}{\theta} \right). \] 
Note that $0 < \Delta_\theta \leq \frac{\klavg{\theta^*}{\theta_0}}{2}$ by Assumption \ref{ass:discrimination}.

To bound the first term on the right of \eqref{eq:rwprobscond}, we introduce the notation $W_\ell(\theta) \equiv W_\ell(\theta, \theta_0, \theta^*, X_\ell, A_\ell) := \log \frac{f_{\theta}(X_\ell|A_\ell)}{f_{\theta_0}(X_\ell|A_\ell)} + \klshort{\theta^*}{\theta}{A_\ell} - \klshort{\theta^*}{\theta_0}{A_\ell}$. With this we can write
\begin{align}
    &\prob{ \psi \geq  \sum_{\ell = 1}^{t-\nu} U_\ell(\theta^*) } = \prob{ \psi \geq  \sum_{\ell = 1}^{t-\nu} E_\ell \left( W_\ell(\theta^*) + \klshort{\theta^*}{\theta_0}{A_\ell}  \right) }  \nonumber \\
    &= \prob{ \sum_{\ell = 1}^{t-\nu} E_\ell  W_\ell(\theta^*) + E_\ell \klshort{\theta^*}{\theta_0}{A_\ell}  < \epsilon (t-\nu) ( \klavg{\theta^*}{\theta_0} - \Delta_\theta) } \nonumber \\
    &\leq \exp \left( - \frac{\epsilon^2 (t-\nu)^2 \Delta_\theta^2}{2(r + D_{\max}^2)(t-\nu)} \right) = \exp \left( - \frac{\epsilon^2 (t-\nu) \Delta_\theta^2}{2(r + D_{\max}^2)} \right), \label{eq:rwprobscond2}
\end{align}
by a standard Chernoff bound for sums of iid subgaussian random variables; this is because each iid random variable $  E_\ell  W_\ell(\theta^*) + E_\ell \klshort{\theta^*}{\theta_0}{A_\ell} $ is subgaussian with (variance) parameter $r + D_{\max}^2$ and mean $\epsilon \klavg{\theta^*}{\theta_0}$.

On the other hand, the second term on the right side of \eqref{eq:rwprobscond} can be bounded as follows:
\begin{align*}
    &\prob{ q + \max_{i=1}^{t-\nu+1} \sum_{\ell = i}^{t-\nu} U_\ell(\theta) \geq  \psi } = \prob{  \max_{i=1}^{t-\nu+1} \sum_{\ell = i}^{t-\nu} E_\ell \left( W_\ell(\theta) + \klshort{\theta^*}{\theta_0}{A_\ell} - \klshort{\theta^*}{\theta}{A_\ell} \right) > \psi - q } \\
    &= \mathbb{P}\left[ \max_{i=1}^{t-\nu+1} \sum_{\ell = i}^{t-\nu} E_\ell  W_\ell(\theta) + \left( E_\ell \klshort{\theta^*}{\theta_0}{A_\ell} - \epsilon \klavg{\theta^*}{\theta_0} \right) + \left( -E_\ell \klshort{\theta^*}{\theta}{A_\ell} +  \epsilon \klavg{\theta^*}{\theta} \right) \right. \\
    &\hspace{2cm}  + \, \epsilon \klavg{\theta^*}{\theta_0}  - \epsilon \klavg{\theta^*}{\theta} > \psi - q \Bigg].
\end{align*}
The first three terms of each summand above are zero mean and subgaussian with a total variance parameter of $r + 2D_{\max}^2$. Denoting their sum by $D_\ell := E_\ell  W_\ell(\theta) + \left( E_\ell \klshort{\theta^*}{\theta_0}{A_\ell} - \epsilon \klavg{\theta^*}{\theta_0} \right) + \left( -E_\ell \klshort{\theta^*}{\theta}{A_\ell} +  \epsilon \klavg{\theta^*}{\theta} \right)$, we  split the analysis into two cases. 

{\em Case 1.} If $\klavg{\theta^*}{\theta_0}  - \klavg{\theta^*}{\theta} > 0$, then 
\begin{align}
    &\prob{  \max_{i=1}^{t-\nu+1} \sum_{\ell = i}^{t-\nu} (D_\ell + \epsilon \klavg{\theta^*}{\theta_0}  - \epsilon \klavg{\theta^*}{\theta}) > \psi - q} \nonumber \\
    &\leq \prob{  \max_{i=1}^{t-\nu+1} \sum_{\ell = i}^{t-\nu} D_\ell + \max_{i=1}^{t-\nu+1} \sum_{\ell = i}^{t-\nu} \epsilon ( \klavg{\theta^*}{\theta_0}  - \klavg{\theta^*}{\theta} ) > \psi - q } \nonumber \\
    &= \prob{  \max_{i=1}^{t-\nu+1} \sum_{\ell = i}^{t-\nu} D_\ell + \epsilon (t-\nu)  ( \klavg{\theta^*}{\theta_0}  - \klavg{\theta^*}{\theta} ) > \epsilon (t-\nu) \left(\klavg{\theta^*}{\theta_0} - \frac{\klavg{\theta^*}{\theta}}{2} \right) - q  } \nonumber \\
    &= \prob{  \max_{i=1}^{t-\nu+1} \sum_{\ell = i}^{t-\nu} D_\ell >  \epsilon (t-\nu) \Delta_\theta - q } \leq \exp \left( - \frac{ \left( \epsilon(t-\nu)  \Delta_\theta-q \right)^2  }{2(r + 2D_{\max}^2)(t-\nu)} \right), \label{eq:rwprobscond3}
\end{align}
thanks to Hoeffding's maximal inequality whenever $\epsilon (t-\nu) \Delta_\theta > q $, see e.g., \cite{jamieson2014lil}.

{\em Case 2.} If $\klavg{\theta^*}{\theta_0}  - \klavg{\theta^*}{\theta} \leq 0$, then 
\begin{align}
    &\prob{  \max_{i=1}^{t-\nu+1} \sum_{\ell = i}^{t-\nu} (D_\ell + \epsilon \klavg{\theta^*}{\theta_0}  - \epsilon \klavg{\theta^*}{\theta}) > \psi - q } \nonumber \\
    &\leq \prob{  \max_{i=1}^{t-\nu+1} \sum_{\ell = i}^{t-\nu} D_\ell + \max_{v=1}^{t-1} \sum_{\ell = v}^{t-1} \epsilon ( \klavg{\theta^*}{\theta_0}  - \klavg{\theta^*}{\theta} ) > \psi - q } \nonumber \\ 
    &= \prob{  \max_{i=1}^{t-\nu+1} \sum_{\ell = i}^{t-\nu} D_\ell >  \epsilon (t-\nu) \Delta_\theta - q } \leq \exp \left( - \frac{ \left( \epsilon(t-\nu)  \Delta_\theta-q \right)^2  }{2(r + 2D_{\max}^2)(t-\nu)} \right), \label{eq:rwprobscond4}
\end{align}
again thanks to Hoeffding's maximal inequality whenever $\epsilon (t-\nu) \Delta_\theta > q $.

Define the following `bad' event $B_t$, representing the situation that the queue $\ExploreQueue_t(\theta^*)$ has not overtaken some other queue $\ExploreQueue_t(\theta)$ by time $t$:
\[ B_t := \bigcup_{\theta : a^*_{\theta} \neq a^*_{\theta^*}} \left\{  \max_{v=1}^{t-1} \log J_{v,t-1}(\theta) \geq \max_{v=1}^{t-1} \log J_{v,t-1}(\theta^*)  \right\}.  \]

Collecting \eqref{eq:expectProb}-\eqref{eq:rwprobscond4}, and  employing an additional union bound over $\ThetaOne$, gives $\forall t \geq \nu$:
\begin{align}
    &\probSup{(\nu, \theta^*)}{ B_t \given \ExploreQueue_\nu } \leq  \nonumber \\
    &\hspace{-0.3cm} \sum_{\theta : a^*_{\theta} \neq a^*_{\theta^*}} \hspace{-5pt}   \left\{
  \exp \left( - \frac{\epsilon^2 (t-\nu) \Delta_\theta^2}{2(r + D_{\max}^2)} \right) +  \indic{\kappa_{t-\nu}(\theta)^c} + \indic{\kappa_{t-\nu}(\theta)}  \exp \left( - \frac{ \left( \epsilon(t-\nu)  \Delta_\theta- \ExploreQueue_\nu(\theta) \right)^2  }{2(t-\nu)(r + 2D_{\max}^2)} \right)   \right\} 
	 \label{eq:timeboundcond0}
\end{align}
where we have defined $\ExploreQueue_\nu \equiv \left( \ExploreQueue_\nu(\theta) \right)_{\theta \in \ThetaOne}$ to be the set of all CUSUM exploration statistics across parameters, at the beginning of round $\nu$, and denoted $\kappa_{t-\nu}(\theta) := \left\{ \ExploreQueue_\nu(\theta) < \epsilon(t-\nu)\Delta_\theta \right\}$.

Note that by the definition of the algorithm, we have $B_t^c \cap \{ E_t = 0 \} \subseteq \{A_t = a^*_{\theta^*} \}$.

{\bf Step 2: Bounding the additional time for the 
 CUSUM queue $\ExploitQueue_t(\theta^*)$ to rise above the threshold and trigger stopping.}

For each $\theta \in \ThetaOne$ and $s, t \in \mathbb{N}$, recall the exploitation-based CUSUM statistic for $\theta_0$ versus $\theta$ \cite{lorden1971procedures}, after having accumulated $t$ rounds worth of observations in exploitation phases: 
\[ \ExploitQueue_{t+1}(\theta) := \log \max_{1 \leq s \leq t+1} \prod_{\ell = s}^{t} \left( \frac{f_{\theta}(X_\ell|A_\ell)}{f_{\theta_0} (X_\ell|A_\ell)} \right)^{1-E_\ell} = \max_{1 \leq s \leq t+1} \sum_{\ell = s}^{t} (1-E_\ell) \log \left( \frac{f_{\theta}(X_\ell|A_\ell)}{f_{\theta_0} (X_\ell|A_\ell)} \right), \]
where the empty product is defined to be $1$, as usual. This statistic satisfies the recursive relation
\[ \ExploitQueue_{t+1}(\theta) = \left( \ExploitQueue_{t}(\theta) + (1-E_t) \log \left( \frac{f_{\theta}(X_t|A_t)}{f_{\theta_0} (X_t|A_t)} \right) \right)^+. \]
Moreover, the algorithm stops as soon as $\max_{\theta \in \ThetaOne} \ExploitQueue_{t}(\theta)$ exceeds the level $\log \beta$. \\

To lighten our notational burden, we henceforth use $\mathbb{P}^{(\nu, \theta^*)}_{Q}$ to denote the conditional measure  $\mathbb{P}^{(\nu, \theta^*)}[\cdot \given \ExploreQueue_\nu]$. We have, for any positive integer $k$, that 
\begin{align}
	\probnew{(\nu, \theta^*)}{\tau \geq \nu + k } &\leq \probnew{\tau \geq \nu +  k, \bigcap_{t=\nu + \frac{k}{2}}^{\nu + k} B_t^c} + \probnew{ \bigcup_{t=\nu + \frac{k}{2}}^{\nu + k} B_t}. \label{eq:timeboundcond1}
\end{align}

Also, by the definition of $\tau$ and the maximum-of-partial-sums property of $\ExploitQueue_t(\theta)$, we have  
\begin{align}
	&\probnew{\tau \geq \nu +  k, \bigcap_{t=\nu + \frac{k}{2}}^{\nu + k} B_t^c} \leq \probnew{ \sum_{\ell=\nu + \frac{k}{2}}^{\nu + k} (1-E_\ell) \log \left( \frac{f_{\theta^*}(X_\ell|A_\ell)}{f_{\theta_0} (X_\ell|A_\ell)} \right) < \log \beta, \bigcap_{t=\nu + \frac{k}{2}}^{\nu + k} B_t^c } \nonumber  \\
	&\leq \probnew{ \sum_{\ell=\nu + \frac{k}{2}}^{\nu + k} (1-E_\ell) \log \left( \frac{f_{\theta^*}(X_\ell|A_\ell)}{f_{\theta_0} (X_\ell|A_\ell)} \right) < \log \beta, \bigcap_{t=\nu + \frac{k}{2}}^{\nu + k} B_t^c, G_{\nu + k/2, \nu + k} } + \probnew{G_{\nu + k/2, \nu + k}^c}, \label{eq:timeboundcond2}
\end{align}
where we have defined the `good' events
\[G_{i,j} := \left\{ \sum_{s=i}^j (1-E_s) \geq \frac{1}{2} (j-i+1) (1-\epsilon)  \right\} \]
for any $i,j$, representing an adequate amount of  exploitation in the time interval $\{i, i+1, \ldots, j\}$. Assuming $\epsilon \leq 1/2$, by Hoeffding's inequality we get 
\begin{equation}
	\probnew{G_{\nu + k/2, \nu + k}^c} \leq e^{-k/16}. \label{eq:timeboundcond3}
\end{equation}

By the law of total probability, we have
\begin{align}
	&\probnew{ \sum_{\ell=\nu + \frac{k}{2}}^{\nu + k} (1-E_\ell) \log \left( \frac{f_{\theta^*}(X_\ell|A_\ell)}{f_{\theta_0} (X_\ell|A_\ell)} \right) < \log \beta, \bigcap_{t=\nu + \frac{k}{2}}^{\nu + k} B_t^c, G_{\nu + k/2, \nu + k} } \nonumber \\
	&= \sum_{j=\frac{1}{4} k (1-\epsilon)}^{k/2} \probnew{\sum_{s=\nu+k/2}^{\nu+k} (1-E_s) = j, \bigcap_{t=\nu + k/2}^{\nu+k} B_t^c} \times \nonumber \\
	& \hspace{1cm} \probnew{\left. \sum_{\ell=\nu+k/2}^{\nu+k} (1-E_\ell) \log \left( \frac{f_{\theta^*}(X_\ell|A_\ell)}{f_{\theta_0} (X_\ell|A_\ell)} \right) < \log \beta \, \right| \sum_{s=\nu+k/2}^{\nu+k} (1-E_s) = j, \bigcap_{t=\nu+k/2}^{\nu+k} B_t^c }. \label{eq:totalprobcond} 
\end{align}

Recall that under $\probnew{\cdot}$, for any $\ell \geq \nu$, the random variable $\log \frac{f_{\theta^*}(X_\ell|A_\ell)}{f_{\theta_0} (X_\ell|A_\ell)}$ is $r$-subgaussian and has mean $\kl{\theta^*(A_\ell)}{\theta_0(A_\ell)}$, conditioned on the past trajectory up to and including $A_\ell$. 
 
Consider now an alternative (and equivalent)   probability space where the sequence of observations from playing the action $a^*_{\theta^*}$ in any exploitation round not earlier than $\nu+ k/2$ (i.e., any round index $\ell \geq \nu+k/2$ with $E_\ell = 0$) is revealed sequentially in order from the iid sequence $Z_1, Z_2, \ldots$, where each $Z_i$ has the probability distribution $\probSup{(1, \theta^*)}{\log \frac{f_{\theta^*}(X|a^*_{\theta^*})}{f_{\theta_0} (X|a^*_{\theta^*})}}$. Define $\mu^* := \kl{\theta^*(a^*_{\theta^*})}{\theta_0(a^*_{\theta^*})}$ to be the mean of each $Z_i$.

We invoke standard subgaussian  concentration of iid sums  in this equivalent probability space, say $\tilde{\mathbb{P}}$, to get
\begin{align*}
	&\probnew{\left. \sum_{\ell=\nu+k/2}^{\nu+k} (1-E_\ell) \log \left( \frac{f_{\theta^*}(X_\ell|A_\ell)}{f_{\theta_0} (X_\ell|A_\ell)} \right) < \log \beta \, \right| \sum_{s=\nu+k/2}^{\nu+k} (1-E_s) = j, \bigcap_{t=\nu+k/2}^{\nu+k} B_t^c } \\
	&= \probEq{\left. \sum_{i=1}^j Z_i < \log \beta \, \right| \sum_{s=\nu+k/2}^{\nu+k} (1-E_s) = j, \bigcap_{t=\nu+k/2}^{\nu+k} B_t^c } \\
	&= \probEq{ \sum_{i=1}^j Z_i < \log \beta } \quad \text{($\because$ $\{Z_i\}_i$ depend only on exploitation outcomes, $\{B_t\}_t$ depend only on exploration outcomes)} \\
	&\leq \exp\left( -\frac{(\log \beta - j\mu^* )^2}{2jr} \right)
\end{align*}
whenever $j \geq \frac{\log 
	\beta}{\mu^*}$, by a Chernoff bound. Using this in \eqref{eq:totalprobcond} gives
\begin{align}
	&\probnew{\sum_{\ell=\nu+k/2}^{\nu+k} (1-E_\ell) \log \left( \frac{f_{\theta^*}(X_\ell|A_\ell)}{f_{\theta_0} (X_\ell|A_\ell)} \right) < \log \beta, \bigcap_{t=\nu+k/2}^{\nu+k} B_t^c, G_{\nu+k/2, \nu+k} } \nonumber \\
	&\leq \max_{j=\frac{1}{4} k (1-\epsilon)}^{k/2} \exp\left( -\frac{(\log \beta - j\mu^* )^2}{2jr} \right) = \exp\left( -\frac{(\log \beta - j^*\mu^* )^2}{2j^*r} \right) \label{eq:timeboundcond4}
\end{align}
with $j^* := \frac{k}{4} (1-\epsilon)$, as long as $j^* \geq \frac{\log \beta}{\mu^*}$ $\Leftrightarrow$ $k \geq \frac{4 \log \beta}{\mu^* (1-\epsilon)}$. This follows by the fact that the function $x \mapsto \frac{x^2}{a+x}$ with $a > 0$  is increasing in $(0, \infty)$. 

{\bf Step 3. Putting together the time bounds to get an overall delay bound.}

Putting together \eqref{eq:timeboundcond0}-\eqref{eq:timeboundcond4} and denoting $\gamma := r+D_{\max}^2$, we get that whenever $k \geq \frac{4 \log \beta}{\mu^* (1-\epsilon)}$,
\begin{align}
	&\probnew{\tau \geq \nu + k} \nonumber \\
	&\leq \hspace{-0.4cm} \sum_{\theta : a^*_{\theta} \neq a^*_{\theta^*}} \sum_{t=\nu+k/2}^{\nu+k} \left\{
  \exp \left( - \frac{\epsilon^2 (t-\nu) \Delta_\theta^2}{2\gamma} \right) + \indic{\kappa_{t-\nu}(\theta)^c} + \indic{\kappa_{t-\nu}(\theta)} e^{ - \frac{ \left( \epsilon(t-\nu)  \Delta_\theta- \ExploreQueue_\nu(\theta) \right)^2  }{2(t-\nu)\gamma} }    \right\} \nonumber  \\
  &\hspace{10pt} + e^{-k/16} + e^{ -\frac{(\log \beta - j^*\mu^* )^2}{2j^*r} } \nonumber \\
    &\leq \hspace{-0.4cm} \sum_{\theta : a^*_{\theta} \neq a^*_{\theta^*}}  \left\{ 
  \frac{ e^{ - \frac{k \epsilon^2 \Delta_\theta^2}{4\gamma} } }{ \left(1 - e^{ - \frac{\epsilon^2 \Delta_\theta^2}{2\gamma} } \right) } +   1 \wedge \sum_{t=\nu+k/2}^{\nu+k}  \left( \indic{\kappa_{t-\nu}(\theta)^c} + \indic{\kappa_{t-\nu}(\theta)} e^{ - \frac{ \left( \epsilon(t-\nu)  \Delta_\theta- \ExploreQueue_\nu(\theta) \right)^2  }{2(t-\nu)\gamma} }   \right)  \right\} \nonumber \\
  &\hspace{10pt} + e^{-k/16} + e^{ -\frac{ \left(\log \beta - \frac{k(1-\epsilon)\mu^*}{4} \right)^2}{\frac{k}{2}(1-\epsilon)r} },\label{eq:tailcond}
\end{align}
where we have denoted $1 \wedge x := \min\{x,1 \}$, and we have taken the minimum of the inner sum (over $t$) with $1$ because probabilities are always bounded by $1$.

The inner sum above for a fixed $\theta$, clamped to $1$, can be bounded as follows. Let $s_0 := \frac{2\ExploreQueue_\nu(\theta)}{\epsilon \Delta_\theta}$, so that $\epsilon s \Delta_\theta - \ExploreQueue_\nu(\theta) \geq \frac{\epsilon s \Delta_\theta}{2}$ whenever $s \geq s_0$. So $\forall \theta$, denoting $\gamma := r + 2D_{\max}^2$, and $1 \wedge x := \min\{x,1 \}$ we have
\begin{align}
      &1 \wedge \sum_{t=\nu+k/2}^{\nu+k}  \left( \indic{\kappa_{t-\nu}(\theta)^c} + \indic{\kappa_{t-\nu}(\theta)} e^{ - \frac{ \left( \epsilon(t-\nu)  \Delta_\theta- \ExploreQueue_\nu(\theta) \right)^2  }{2(t-\nu)\gamma} } \right) \nonumber \\
      &= 1 \wedge \sum_{s=k/2}^{k} \left( \indic{\kappa_s(\theta)^c} + \indic{\kappa_s(\theta)} e^{ - \frac{ \left( \epsilon s  \Delta_\theta- \ExploreQueue_\nu(\theta) \right)^2  }{2s\gamma} } \right) \nonumber \\
      &\leq \indic{k/2 \geq s_0} \sum_{s=k/2}^{k} e^{ - \frac{  \epsilon^2 s  \Delta_\theta^2   }{8\gamma} } + \indic{k/2 < s_0}  \cdot 1 \nonumber \\
      &\leq \indic{k \geq \frac{4\ExploreQueue_\nu(\theta)}{\epsilon \Delta_\theta}} \left(\frac{e^{-\frac{  k \epsilon^2   \Delta_\theta^2 }{16\gamma} } }{1- e^{- \frac{  \epsilon^2   \Delta_\theta^2   }{8\gamma} } } \right) + \indic{k < \frac{4\ExploreQueue_\nu(\theta)}{\epsilon \Delta_\theta} }  \cdot 1. \label{eq:tailcond2} 
\end{align}

We are now in a position to obtain a bound on the (conditional) expected excess detection delay $\expectSup{(\nu, \theta^*)}{(\tau-\nu)^+}$ by integrating the tail  and using \eqref{eq:tailcond} and \eqref{eq:tailcond2}:
\begin{align*}
    &\expectnew{(\tau-\nu)^+} = \sum_{k=1}^\infty \probnew{(\tau-\nu)^+ \geq k} = \sum_{k=1}^\infty \probnew{\tau \geq \nu + k} \\
    &\leq 20 + \frac{4 \log \beta}{\mu^* (1-\epsilon)} + \sum_{k=\left\lceil \frac{4 \log \beta}{\mu^* (1-\epsilon)} \right\rceil }^\infty e^{ -\frac{ \left(\log \beta - \frac{k(1-\epsilon)\mu^*}{4} \right)^2}{\frac{k}{2}(1-\epsilon)r} } \\
    &+    \sum_{\theta : a^*_{\theta} \neq a^*_{\theta^*}} \left\{   
  \sum_{k=\left\lceil \frac{4 \log \beta}{\mu^* (1-\epsilon)} \right\rceil}^\infty \frac{ e^{ - \frac{k \epsilon^2 \Delta_\theta^2}{4\gamma} } }{ \left(1 - e^{ - \frac{\epsilon^2 \Delta_\theta^2}{2\gamma} } \right) } + \frac{4\ExploreQueue_\nu(\theta)}{\epsilon \Delta_\theta} +  \sum_{k=\left\lceil \frac{4 \log \beta}{\mu^* (1-\epsilon)} \right\rceil}^\infty  \frac{e^{-\frac{  k \epsilon^2   \Delta_\theta^2 }{16\gamma} } }{ \left( 1- e^{- \frac{  \epsilon^2   \Delta_\theta^2   }{8\gamma} } \right) } \right\}.
\end{align*}

The third term above admits the bound
\begin{align*}
	&\sum_{k=\left\lceil \frac{4 \log \beta}{\mu^* (1-\epsilon)} \right\rceil }^\infty e^{ -\frac{ \left(\log \beta - \frac{k(1-\epsilon)\mu^*}{4} \right)^2}{\frac{k}{2}(1-\epsilon)r} } = \sum_{k=\left\lceil \frac{4 \log \beta}{\mu^* (1-\epsilon)} \right\rceil }^{\left\lceil \frac{8 \log \beta}{\mu^* (1-\epsilon)} \right\rceil} e^{ -\frac{ \left(\log \beta - \frac{k(1-\epsilon)\mu^*}{4} \right)^2}{\frac{k}{2}(1-\epsilon)r} } + \sum_{k=\left\lceil \frac{8 \log \beta}{\mu^* (1-\epsilon)} \right\rceil }^\infty e^{ -\frac{ \left(\log \beta - \frac{k(1-\epsilon)\mu^*}{4} \right)^2}{\frac{k}{2}(1-\epsilon)r} } \\
	&\leq 1 + \frac{4 \log \beta}{\mu^* (1-\epsilon)} + \frac{e^{-\frac{\mu^* \log \beta}{4r}}}{1-e^{-\frac{(\mu^*)^2(1-\epsilon)}{32r}}},
\end{align*}
while each summand corresponding to $\theta$ in the final term is bounded as
\begin{align*}
	&\sum_{k=\left\lceil \frac{4 \log \beta}{\mu^* (1-\epsilon)} \right\rceil}^\infty \frac{ e^{ - \frac{k \epsilon^2 \Delta_\theta^2}{4\gamma} } }{ \left(1 - e^{ - \frac{\epsilon^2 \Delta_\theta^2}{2\gamma} } \right) } + \frac{4\ExploreQueue_\nu(\theta)}{\epsilon \Delta_\theta} +  \sum_{k=\left\lceil \frac{4 \log \beta}{\mu^* (1-\epsilon)} \right\rceil}^\infty  \frac{e^{-\frac{  k \epsilon^2   \Delta_\theta^2 }{16\gamma} } }{ \left( 1- e^{- \frac{  \epsilon^2   \Delta_\theta^2   }{8\gamma} } \right) } \leq \frac{4\ExploreQueue_\nu(\theta)}{\epsilon \Delta_\theta} + \frac{2e^{-\frac{\epsilon^2\Delta_\theta^2 \log \beta}{4\gamma\mu^*(1-\epsilon)}}}{1-e^{-\frac{\epsilon^2 \Delta_\theta^2}{8\gamma}}},
\end{align*}
giving 
\begin{align*}
	\expectnew{(\tau-\nu)^+} &\leq 21 + \frac{8 \log \beta}{\mu^* (1-\epsilon)} +  \sum_{\theta : a^*_{\theta} \neq a^*_{\theta^*}} \left( \frac{4\ExploreQueue_\nu(\theta)}{\epsilon \Delta_\theta} + \frac{2e^{-\frac{\epsilon^2\Delta_\theta^2 \log \beta}{4\gamma\mu^*(1-\epsilon)}}}{1-e^{-\frac{\epsilon^2 \Delta_\theta^2}{8\gamma}}} \right),
\end{align*}
where (recall) $\mu^* := \kl{\theta^*(a^*_{\theta^*})}{\theta_0(a^*_{\theta^*})}$ and $\gamma = r + 2D_{\max}^2$.

Taking expectation under the $\mathbb{P}^{(\infty)}$ distribution of $Q_\nu$ completes the proof of Theorem \ref{thm:epsgreedygenchange}.

\begin{lemma}
\label{lem:partialsums}
For any sequence $x_1, \ldots, x_{t-1}$ and $\nu \in [t]$, 
\[ \max_{i=1}^t \sum_{\ell = i}^{t-1} x_\ell \leq \left( \max_{i=1}^{\nu - 1} \sum_{\ell = i}^{\nu - 1} x_\ell \right)^+ + \max_{i=\nu}^{t} \sum_{\ell = i}^{t-1} x_\ell =  \max_{i=1}^{\nu} \sum_{\ell = i}^{\nu - 1} x_\ell  + \max_{i=\nu}^{t} \sum_{\ell = i}^{t-1} x_\ell. \]
\end{lemma}
\begin{proof}[Proof of Lemma \ref{lem:partialsums}]
	The lemma is a consequence of the  elementary result that $\max_i (a_i + b_i) \leq \max_i a_i + \max_i b_i$ for any discrete collection of numbers $\{a_i\}_i$, $\{b_i\}_i$.
\end{proof}

\section{Experiment Details}
This section describes in detail the setup and methodology followed for obtaining the results in Section \ref{sec:expts}. It also includes additional results, both for the synthetic and audio sensing settings, that explore the impact of various problem parameters on performance. 

\subsection{Version of the \egcd\, algorithm used in experiments}

In all our experiments, we used the {\em full data-MLE} implementation of the \egcd\, template, as given in Algorithm \ref{algo:epsgreedy-fullMLE}. The only difference of this algorithm from the {\em exploration data-MLE} version given in the main text (Algorithm \ref{algo:epsgreedy}) is that the estimate $\hat{\theta}_t$ for the post-change distribution is computed using data from {\em all} previous rounds, regardless of exploration or exploitation.

\begin{algorithm}
	\renewcommand\thealgorithm{1}
	\caption{\egcd\, (Full data-MLE version)}\label{algo:epsgreedy-fullMLE}
	\begin{algorithmic}[1]
		\STATE \textbf{Input:}  Exploration rate $\epsilon \in [0,1]$, Pre-change parameter $\theta_0$, Post-change parameter set  $\Theta^{(1)}$, Stopping threshold $\beta > 0$, Exploration distribution over actions $\pi$, Observation function $g_\theta(x, a)$ $\forall (\theta, x, a) \in \ThetaOne \times \mathcal{X} \times \mathcal{A}$.
		
		\STATE {\bf Init:} $\AlldataQueue_1(\theta) \leftarrow 0$, $\ExploitQueue_1(\theta) \leftarrow 0$ $\forall \theta \in \ThetaOne$ \hfill \COMMENT{CUSUM statistics based on overall and exploitation-only data, per parameter}
		
		\FOR{round $t = 1, 2, 3, \ldots$}
		\IF{$\max_{\theta \in \ThetaOne} \label{line:stopping} \ExploitQueue_t(\theta) \geq \beta$}
		\STATE {\bf break} \hfill \COMMENT{Stop sampling and exit}
		\ENDIF
		\STATE Sample $E_t \sim \text{Ber}(\epsilon)$ independently
		\IF{$E_t == 1$} 
		\STATE Play action $A_t \sim \pi$ independently  \hfill \COMMENT{Explore}
		\STATE Get observation $X_t$
		
		\ELSE
		\STATE Compute $\hat{\theta}_t = \argmax_{\theta \in \Theta^{(1)}} \ExploreQueue_t(\theta)    $ \hfill\COMMENT{Most likely post-change distribution based on all past data} \label{line:MLE}
		\STATE Play action $A_t = \argmax_{a \in \cal{A}} \kl{\theta^{(0)}(a)}{\hat{\theta}_t(a)}$
		\STATE Get observation $X_t$
		\STATE Set $\forall \theta \in \ThetaOne:$ $\ExploitQueue_{t+1}(\theta) \leftarrow \left( \ExploitQueue_t(\theta) + g_\theta(X_t, A_t) \right)^+$ \hfill \COMMENT{Update exploitation CUSUM statistics}
		\ENDIF
		\STATE Set $\forall \theta \in \ThetaOne:$ $\AlldataQueue_{t+1}(\theta) \leftarrow \left( \AlldataQueue_t(\theta) + g_\theta(X_t, A_t) \right)^+$ \hfill \COMMENT{Update overall data CUSUM statistics}
		\ENDFOR
	\end{algorithmic}
\end{algorithm}

The chief reason to prefer the exploration-only version in the main text is that the theoretical analysis of its detection delay is slightly simpler than the full-data version. This is because the  exploration-only CUSUM statistic  (queue) changes from one time to the next in an essentially memoryless manner since the sensing action is chosen independent of the past. 

On the other hand, we preferred the full data-version in experiments since it was noticed to offer slightly better numerical performance. We remark that the full-data MLE version can also be analyzed rigorously for its detection delay\footnote{The false alarm rate analysis remains the same as there is no change to the stopping criterion in the algorithm.} in a manner similar to that of Algorithm \ref{algo:epsgreedy}, with an essentially similar delay guarantee. However, one will have to content with an extra overhead in the additive term of the detection delay (i.e., the last term of \eqref{eq:detdelaybound}), due to not being able to apply time-uniform maximal inequalities for martingales (see \eqref{eq:rwprobscond3}, \eqref{eq:rwprobscond4} in Section \ref{app:detectiondelay}) but resort to a slightly worse union bound over time.

\subsection{Algorithms compared in the experiments}

We compare the performance of the following adaptive sensing change detection algorithms in all our experimental settings: 

\begin{enumerate}
	\item The \egcd\, algorithm (`EG' in plots), used with $\epsilon = 0.2$
    \item Uniform Random Sampling (`URS'): The non-adaptive sensing rule that plays an action drawn uniformly at random from a given set of actions; this is \egcd\, with $\epsilon = 0$
    \item Oracle Sampling (`Oracle'): The sensing rule that always plays the most informative  action knowing the post-change distribution in advance:  $\argmax_{a \in \mathcal{A}}  \kl{\theta^*(a)}{\theta^{(0)}(a)}$.
    
\end{enumerate}

\subsection{Synthetic experiments}

We conduct change detection experiments on a line graph, of size $N$, serving as an ambient space. Nodes of the graph are interpreted as physical locations, and take values in $[N]$. Nodes $j,k$ are connected if $|j-k|=1$. Each node $n \in [N]$ offers a Gaussian-distributed observation depending on the changepoint $\nu \in \mathbb{N}$. In particular, the signal (observations from all nodes) at time $t$ is a random vector $S(t) = (S_n(t))_{n \in [N]} \in \mathbb{R}^N$, where 
\begin{equation} \label{eq.signal}
S_n(t) = \theta_n \indic{t \geq \nu}+W_n(t),\,\,t=0,1, 2, \ldots    
\end{equation}
$\theta := (\theta_n)_{n \in [N]}$ represents the post-change parameter and $W_n(t) \in {\cal N}(0,\sigma^2)$ are IID Gaussian random variables for $n \in [N]$ and across time $t=0,1,\ldots,$, and represents observation noise. We also choose $\sigma^2=1/2$ for all our synthetic experiments. Note that in essence, this setup has the pre-change parameter set to zero ($\theta=0 \in \mathbb{R}^N$).  


{\it Isolated and Structured Anomalies.} We consider two types of the vector of change parameters: \\(a) Isolated singleton change, namely, $\theta_n \in \{0,\,1\}$ and $\sum_{n \in [N]}\theta_n=1$; \\(b) Structured K-change: We consider parameter changes with\footnote{We define $\mbox{Supp}(x) := \{i: x_i \neq 0\}$.} $|\mbox{Supp}(\theta)|=K$, and the nodes (components) corresponding to the non-zero support are connected. As such the collection of anomalies is $N-K+1$ corresponding to different starting positions.

{\it Diffuse and `Pointy' Action sets.} In a parallel fashion we allow actions to be vertices of the $N$-hypercube, $a_n \in {\cal A} \subset \{0,\,1\}^N$, and the action sets to be either {\em pointy} (${\cal A}_1$), namely, $|\mbox{Supp}(a)|=1$ $\forall a \in {\cal A}_1$, which allows probing only single nodes, or {\em diffuse} (${\cal A}_2$), where only a connected subset of nodes can be queried. In either case, the observation received on an action, $a \in {\cal A}$ is given by $X_a = \langle \frac{a}{\|a\|_2},S \rangle$, where we impose the normalization because we want to maintain the same signal-to-noise ratio (SNR) across different types of probes.

We reported results for pointy actions and isolated anomalies in Sec.~\ref{sec:expts}. We will describe experiments with other scenarios here. 

{\bf Structured Anomalies and Diffuse Action Sets.}\\
We experiment with Structured K-change as described above with $K=5$. The action sets are diffuse: $${\cal A}=\{a \in \{0,1\}^N: \|a\|_2=1, |\mbox{Supp}(a)|=5, \,\, \mbox{$a$ is connected}.\}$$ 
This means that corresponding to each anomalous change, there is an action (unknown to the learner) that perfectly overlaps with the entire anomalous change. Furthermore, there are several other actions that partially overlap with the structured anomaly. As a result there is a higher probability of detecting anomalies. 

As in Sec.~\ref{sec:expts} we tabulate results for change point $\nu=40$, for different sizes of graph. Our results are based on 5000 Monte Carlo runs. The mean and standard deviation for change point $\nu=40$ is reported in Table~\ref{tab.SAPA}. We see that, although diffuse, such actions appear to improve detection delay for \egcd\, in comparison to the case considered in Sec.~\ref{sec:expts}. This is to be expected because the number of structured anomalies are smaller. For instance for a graph of size $10$, we only have $5$ anomalous parameter changes. Furthermore, we can detect anomalies even when the actions only partially overlap with the anomaly. Thus change detection methods now have a larger probability to detect parameter changes in contrast to isolated anomalies.

\begin{table}[h!]
\begin{center}
\begin{tabular}{ccccc}
\toprule  
 Size    & Oracle & \egcd(Full) &  \egcd(Theory) & URS \\
 \midrule
 10  & $51\pm7  $ & $64\pm12 $ & $90\pm21 $ & $96\pm14 $ \\  \midrule
 15   & $51\pm7 $ & $71\pm13 $ & $100\pm27  $ & $163 \pm26  $\\  \midrule
 20    & $51\pm7  $ & $74\pm14$ & $113\pm35 $ & $237 \pm40 $\\  \midrule
 25   & $51\pm7  $ & $80\pm17 $ & $126\pm46 $ & $310\pm52 $ \\  \midrule
\end{tabular}
\end{center}
\caption{Structured Anomalies with Diffuse Actions: Observed mean and standard deviation for the simulated stopped time for varying graph-sizes with diffused action set for change occurring at $\nu=40$.}\label{tab.SAPA}
\end{table}
We also observe that \egcd\, is still as effective, and closely mirrors Oracle performance. We do not tabulate the effect of different changepoints here. This is because, we notice that the changepoint parameter $\nu$ has no noticeable effect when graph size is held constant for all of the reported methods (Oracle, URS, and \egcd\,). 

{\bf Structured Anomalies with Pointy Action Sets.}\\
Here we experiment with $K=5$ as in the setup above but examine the effect of pointy action sets. As a result our actions can only probe some component of the anomaly. Observe that the anomaly is spread across a larger region. Therefore, a pointy anomaly can only sample a small part of the parameter change in any round.

We report mean and variance for expected delay for change point $\nu=40$, for different graph sizes over 5000 Monte Carlo runs in Table~\ref{tab.SADA}.
\begin{table}[h!]
\begin{center}
\begin{tabular}{ccccc}
\toprule  
 Size    & Oracle & \egcd(Full) &  \egcd(Theory) & URS \\
 \midrule
 10  & $150\pm27  $ & $192\pm41 $ & $289\pm124 $ & $299\pm57 $ \\  \midrule
 15   & $149\pm27 $ & $210\pm49 $ & $340\pm175  $ & $448 \pm85  $\\  \midrule
 20    &$150\pm27  $ & $204\pm44$ & $350\pm213 $ & $597\pm118 $\\  \midrule
 25   & $149\pm27  $ & $213\pm48$ & $395\pm246 $ & $746\pm145 $\\  \midrule
\end{tabular}
\end{center}
\caption{Structured Anomalies with Pointy Action Sets. Observed mean and standard deviation for the simulated stopped time for varying graph-sizes at $\nu=40$.}
\label{tab.SADA}
\end{table}

In this experiment both Oracle and \egcd\, exhibit larger delays. The reason now is that pointy anomalies can only sample a single component, and as such a component in the anomalous region exhibits smaller change, and so it takes a longer time to detect. Again, no noticeable impact of varying changepoint on delay was observed.

{\bf Isolated Anomalies with Diffuse Action Sets.}\\
Here we consider the case where the anomalies are isolated but the action sets are diffuse. Our results (mean and variance) over 5000 Monte Carlo runs for changepoint $\nu=40$ is tabulated in Table~\ref{tab.IADA}.

\begin{table}[h!]
\begin{center}
\begin{tabular}{ccccc}
\toprule  
 Size    & Oracle & \egcd+ &  \egcd(Theory) & URS \\
 \midrule
 10  & $249\pm35 $ & $277\pm 40$ & $321\pm 49$ & $497\pm 74$ \\  \midrule
 15   & $250\pm35 $ & $281\pm 40$ & $326\pm 54 $ & $ 549\pm 81 $\\  \midrule
 20    &$249\pm35 $ & $297\pm42$ & $370\pm84 $ & $998 \pm151 $\\  \midrule
 25   & $249\pm35 $ & $324\pm52$ & $509\pm237 $ & $1044 \pm157 $\\  \midrule
\end{tabular}
\end{center}
\caption{Isolated Anomalies with Diffuse Action Sets. Observed mean and standard deviation for the simulated stopped time for varying graph-sizes for change occurring at $\nu=40$.}
\label{tab.IADA}
\end{table}

Among all of the different scenarios, this setup has uniformly larger expected delay across all of the methods (Oracle, \egcd\, and URS). This is not surprising considering the fact that isolated anomalies when probed with diffuse actions manifest as substantially smaller change. This is because a diffuse action, spread across 5 locations, is capable of collecting only a 5th of the energy of the anomaly. 

\section{Experiment Details for Real-World Audio Dataset}
\label{app:audioexpts}
Recall from Sec.~\ref{sec:expts} we explored changepoint detection on the MMII dataset. We pointed out that we used reconstruction errors of auto-encoders, and modeled these errors with Gaussians. Here we provide more details and additional experiments on the dataset.

\textit{Audio Processing}. For each audio-stream we train auto-encoders on normal data using mel-spectrogram features. We train different autoencoders for different machine ids. We use mean of reconstruction errors from each of these machines when there are no anomalies, and construct the  pre-change parameter vector. Similarly, we use mean of reconstruction errors from each of these machines when there are anomalies and construct post-change parameters.  

To compute reconstruction errors, we adopt the autoencoder architecture as used in Section 4 in \cite{purohit_harsh_2019_3384388}. We also make use of publicly available \href{https://github.com/aws-samples/sound-anomaly-detection-for-manufacturing}{code} to train autoencoders with mel-spectrogram features of normal data as inputs. We use the same parameters that are used in ~\cite{purohit_harsh_2019_3384388}, to extract mel-spectrogram features from a given audio input.  We assume that, for a given audio stream, there is only one anomaly, and that anomaly is present in only one of the machine ids. 

The resulting pre-change parameters across the 4 machines and the post-change parameters under an anomaly are displayed below:

\begin{table}[h!]
  \centering
  \begin{tabular}{|c|c|c|}
    \hline
    \textbf{Machine ID} & \multicolumn{2}{c|}{\textbf{Mean reconstruction error}} \\
    \cline{2-3}
    & \textbf{Normal} & \textbf{Abnormal} \\
    \hline
    00 & 7.816003 & 18.043417 \\ \hline
    02 & 7.728631 & 12.879204 \\ \hline
    04 & 12.029381 & 15.425252  \\ \hline
    06 & 9.34813 & 10.788003   \\ \hline
  \end{tabular}
  \caption{\footnotesize Mean reconstruction errors for machine ID $00,02,04,06$ under normal and abnormal operation}
\end{table}

Using the notation of our synthetic experiment, our setup here can be described as the case with isolated anomalies (i.e., only one machine has an anomaly), and pointy actions (i.e., we can only query one machine at any time). 

\begin{figure}[h!]
    \centering
    \includegraphics[width=0.8\textwidth]{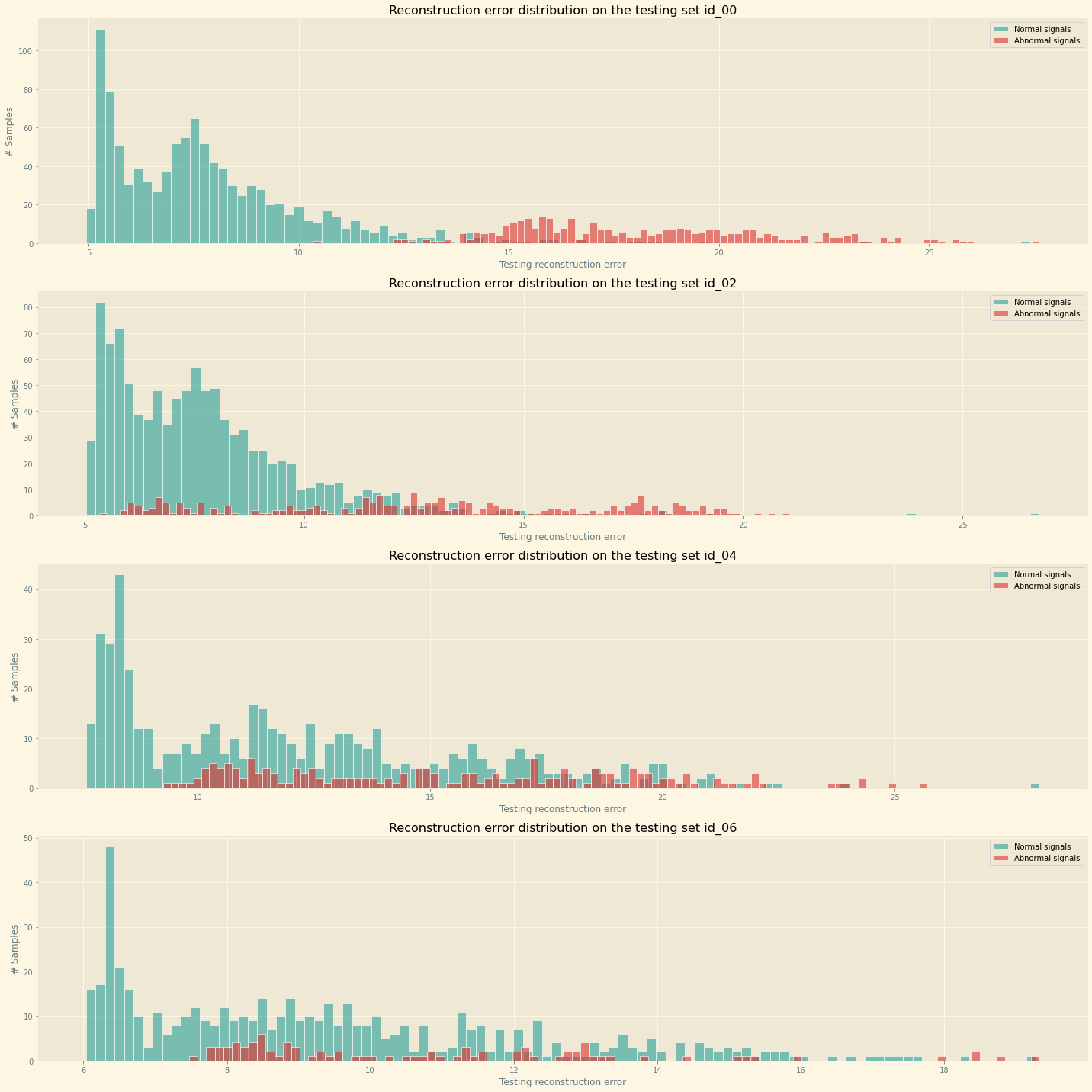}
    \caption{Histogram of reconstruction errors under normal and abnormal operation of machine id $00,02,04,06$. }
\vspace{-6pt}
\end{figure}

\begin{figure}[h!]
    \centering
    \includegraphics[width=0.8\textwidth]{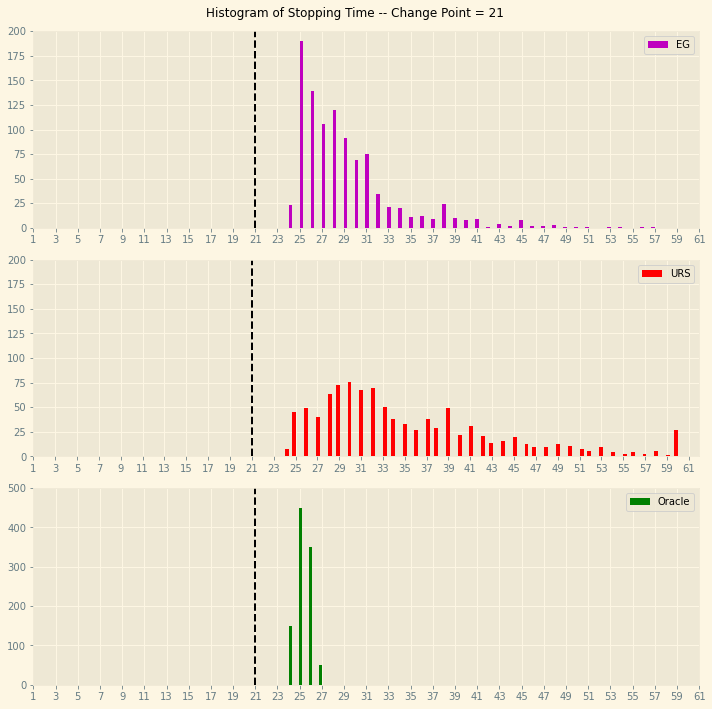}
    \caption{Audio based change detection of machine anomaly: Histogram of stopping times by URS, Oracle, and \egcd\, (EG) for changepoint $\nu=21$.}
    \label{19}
\vspace{-10pt}
\end{figure}

{\it Experiment.} 
%
In addition to the $\nu=6$ case, which we reported in the main paper, we simulate
changepoints for $\nu=21$. This corresponds to $210$ seconds. We do this by introducing anomalies in machine bearing ID00 as follows. We concatenated 21 normal files and 39 abnormal files chosen uniformly at random from machine ID00. For the other machines we concatenated 60 normal files at random. The 60 files correspond to 600 seconds. Our changepoint corresponds to 21st file, which we denote as $\nu=21$ and our task is to detect this change. Note that both the machine ID and the changepoint is not known to the learner. Our results are depicted as histograms for changepoints of anomaly detection in Fig.~\ref{19}.

As observed Oracle method has a small variance, and the histogram is concentrated at around $25$, which is about $40$ seconds delay. \egcd\, also exhibits small delay, but its variance is somewhat larger in this context. URS expected delay and variance are substantially larger. This demostrates the gains due to adaptive processing.

\section{Detailed Survey of Related Work}
\label{app:relatedwork}
Changepoint detection deals with identifying points in time when probability laws governing a stochastic process changes abruptly. The problem of changepoint detection has been widely studied, dating back to the pioneering works of Page~\cite{page} and Lorden \cite{lorden1971procedures}.

Online change detection focuses on situations where the data is obtained incrementally over time, and one must infer whether a change has occurred at each time. A large part of the online change detection literature, like our paper, adopts a frequentist approach, and, in particular, utilizes parametric models for pre-change and post-change distributions. In this context, the CUSUM algorithm \cite{page} and its variants such as the generalized likelihood ratio statistics have been proposed, in which a change is announced when the likelihood ratio statistic exceeds a threshold. While there are a number of prior works on this topic (see ~\cite{Basseville,Chen12,veeravalli2014quickest}), specific attention to finite time (i.e., non-asymptotic) guarantees on detection delay is more recent~\cite{tllai10,maillard19a}, and as such remains somewhat open (\cite{Aminikhanghahi17}). While changepoint detection has also been studied from a Bayesian perspective, much of this literature has focused on the batch setting (see \cite{gundersen2021active,adams2007bayesian,Fearnhead_2007}). Though recent works have begun to focus on Bayesian online detection, there has been little work (apart from \cite{alami20a}) on proving finite time guarantees.

Our work is motivated by costs imposed on data collection due to resource constraints. In this context, while adopting a frequentist perspective, we propose methods for adaptive online data selection for multi-stream time-series data. Recent works have begun to focus attention on adaptive online data collection for changepoint detection from both frequentist~\cite{zhang2020bandit} and Bayesian~\cite{osborne,gundersen2021active} perspectives. Additionally, while different from our focus, we mention in passing that methods for active change-point detection~\cite{hayashi19}, where the task is to adaptively determine the next input have also been proposed. Furthermore, there are a number of works that focus on bandit regret minimization for non-stationary time-series data~\cite{garivier2011upper, liu2018change, mellor2013thompson}. While these works are related we note that regret minimization 
is a fundamentally different objective from changepoint identification where the goal is to minimize detection delay.  

We will outline similarities and differences between our work and closely related prior works. 
From a practical perspective, \cite{osborne} is similar to our work in that they too motivate their approach from a sensor network viewpoint, where sensors may undergo faults or changepoints exhibited due to environmental factors. They propose a Bayesian formalism largely based on the well-known Bayesian online change detection (BOCD) method~\cite{adams2007bayesian}, and leverage Gaussian processes (GP) for modeling. The GP perspective allows for tractable sequential time-series prediction and sensor selection. Recently, \cite{gundersen2021active} have proposed to extend the BOCD approach to incorporate costs in making decisions for real-time data acquisition in multi-fidelity sensing scenarios. Different from these perspectives, our approach is frequentist, and does not impose action-specific costs and distributional constraints on the underlying latent parameters or on the changepoints. Furthermore, we derive finite time performance bounds, which is not a focus of any of these works. 

Similar to our work,  \cite{zhang2020bandit} also adopts a frequentist perspective and proposes a method for bandit changepoint detection based on the well-known Shiryaev-Roberts-Pollak scheme. To adaptively choose data streams/sensors, they utilize Thompson sampling \cite{thompson1933likelihood} to balance exploration of different data streams for acquiring knowledge with exploitation of informative streams. In addition, they present theoretical properties, and show that their method does not trigger false alarm too soon. However, detection delays under false alarm constraints are not explicitly characterized, which is a key challenge.

In summary, the principal difference between these prior works on bandit online changepoint detection and our \egcd\, is that we are able to explicitly characterize information theoretic lower bounds on expected detection delay under a false alarm constraint. Furthermore \egcd\, is natural variant of CUSUM, and our explicit analysis shows that it exhibits finite-time performance guarantees on expected detection delay and matches our lower bound at low false alarm rates. This leads to optimality that is absent in prior work.

\section{Discussion and Future Work}

This work has laid down a principled approach to exploration with information-limited sensing to rapidly detect changes in distribution. Specifically, we have shown that relatively `simple' (epsilon-greedy) forced exploration is sufficient to obtain detection delays comparable to an oracle who knows the post-change distribution beforehand. 

As such, this study represents only an initial attempt to understand the limits of adaptive sensing for change detection, and opens up a host of interesting avenues for further investigation. These include (a) the possibility of `more adaptive' exploration approaches, such as confidence-set or posterior sampling-based methods, that could improve the delay for learning a good guess of the post-change distribution (the second term in the detection delay bound), (b) adaptive sensing when both the pre change and post change distribution is unknown, which also entails learning the default distribution online, (c) extensions to continuous parameter spaces, (d) detecting multiple changes that occur continually over time, and (e) studying the adaptive change detection problem for Markovian dynamics or controlled processes.
\end{document}